\documentclass[journal]{IEEEtran}
\usepackage[ruled,linesnumbered]{algorithm2e}
\usepackage{algpseudocode}
\usepackage{listings}

\usepackage{float}
\usepackage{graphicx}
\usepackage{multirow}
\usepackage{amsmath} % assumes amsmath package installed
\usepackage{amssymb}  % assumes amsmath package installed
\usepackage[normalem]{ulem}
\usepackage{booktabs}
\usepackage{footnote}
\usepackage{threeparttable}
\usepackage[colorlinks,linkcolor=red]{hyperref}
\usepackage{cite}
\usepackage{color}
\usepackage{booktabs}
\usepackage[table]{xcolor}
\usepackage{caption}
\usepackage{float} 
\usepackage{subcaption}

\usepackage{xcolor}
\definecolor{mycolorpurple}{RGB}{255, 102, 255}
\definecolor{mycolorred}{RGB}{227, 26, 26}
\definecolor{mycolorskyblue}{RGB}{6, 185, 238}

\begin{document}

\title{\textcolor{black}{SparseDet}: A Simple and Effective Framework for Fully Sparse LiDAR-based 3D Object Detection}

\author{ Lin Liu, Ziying Song, Qiming Xia, Feiyang Jia, Caiyan Jia, Lei Yang, Hongyu Pan

\thanks{This work was supported by the National Key R\&D Program of China (2018AAA0100302).\emph{(Corresponding author: Caiyan Jia.)}}

\thanks{Lin Liu, Ziying Song, Feiyang Jia, Caiyan Jia are with School of Computer and Information Technology, Beijing Key Lab of Traffic Data Analysis and Mining, Beijing Jiaotong University, Beijing 100044, China (e-mail: 22110110@bjtu.edu.cn; liulin010811@gmail.com; cyjia@bjtu.edu.cn.)}

\thanks{Qiming Xia is with  Fujian Key Laboratory of Sensing and Computing for Smart Cities, Xiamen University, Xiamen, China, Fujian 100044, China (e-mail: 22110110@bjtu.edu.cn; liulin010811@gmail.com; cyjia@bjtu.edu.cn.)}

\thanks{Lei Yang is with the State Key Laboratory of Intelligent Green Vehicle and Mobility, and the School of Vehicle and Mobility, Tsinghua University, Beijing 100084, China (e-mail: yanglei20@mails.tsinghua.edu.cn).}

\thanks{Hongyu Pan is with Horizon Robotics (e-mail: karry.pan@horizon.cc).}

}

% \thanks{Jun Xie, Feng Chen are with Lenovo Research, Beijing 100085, China (xiejun@lenovo.com, chenfeng13@lenovo.com)}

% SparseDet 与其他稀疏检测方法的对比，(a) 该系列的代表方法VoxelNext 和 SAFDNet通过特征扩散用堆栈卷积层填充其中心体素，然而，仅使用单个中心体素特征作为物体的代理大量的实例特征未被利用，削弱了以中心点呈现物体的能力。(b) 该系列的代表方法FSD 和FSDV2 利用投票机制，将前景点聚集成以对象为中心的集群，以进行进一步的预测。然而，该系列方法过于依赖于点分割和预测细化。由于需要调优的大量超参数。导致时间延迟的问题。(c) Our SparseDet 将稀疏查询作为物体的候选，并选择性地聚合感兴趣位置的稀疏体素特征，避免了中心体素特征信息表达能力不足的问题。同时，SparseDet不需要额外的任务进行辅助。
\maketitle
% 摘要：稀疏检测方案不依赖密集特征图从而避免了随检测距离二次增长的计算成本，从而具备扩展到远程检测的能力。近年来受到越来越多的关注。一些方法通过利用堆栈卷积层将稀疏特征扩散至中心体素出，尽管了实现高效的推理速度却削弱了中心点呈现物体的能力，导致模型表现出较差的性能。Others methods 利用投票机制，将前景点聚集成以对象为中心的集群，以进行进一步的预测。尽管充分聚合了上下文信息，然而却过于依赖额外的辅助任务导致推理速度不佳。In this paper, 我们设计了SparseDet, a  simple and effective end-to-end framework for fully sparse 3D Object Detection. 具体地，我们将稀疏查询设计为物体的代理避免了中心特征缺失的问题, 同时为了充分捕获点云的上下文信息来增强物体代理呈现物体的能力。我们设计了两个关键模块：局部多尺度特征聚合模块 (which 仅通过利用简单地坐标转换和最邻近关系实现稀疏关键体素在不同尺度上的特征融合，从而来捕捉点云中的细节信息)和全局特征聚合模块通过选择性聚合全局上下文信息来提升模型对小目标和远距离物体检测的性能。基于nuS数据集上的实验表明了我们提出的方法的有效性。在1/4 split 上我们的方法在保证推理速度的同时，相比基线模型voxelnext在Motor,Bike,Car 类别上mAP分别提升了5.57%,3.60% ,2.19%.
% 刘林修改：基于激光雷达的稀疏三维物体检测由于计算效率的优势在自动驾驶中发挥着至关重要的作用。现有的方法要么将单个中心体素特征作为物体代理，要么将前景点聚合的集群视为物体代理。然而，前者缺乏上下文聚合能力导致物体代理的信息表达能力不足进而性能次优。后者则过于依赖多阶段管道和辅助任务降低其推理速度上的优势。

\begin{abstract}
LiDAR-based sparse 3D object detection plays a crucial role in autonomous driving applications due to its computational efficiency advantages. Existing methods either use the features of a single central voxel as an object proxy, or treat an aggregated cluster of foreground points as an object proxy. However, the former lacks the ability to aggregate contextual information, resulting in insufficient information expression in object proxies. The latter relies on multi-stage pipelines and auxiliary tasks, which reduce the inference speed. To maintain the efficiency of the sparse framework while fully aggregating contextual information, in this work, we propose SparseDet which designs sparse queries as object proxies. It introduces two key modules, the Local Multi-scale Feature Aggregation (LMFA) module and the Global Feature Aggregation (GFA) module, aiming to fully capture the contextual information, thereby enhancing the ability of the proxies to represent objects. Where LMFA sub-module achieves feature fusion across different scales for sparse key voxels %which does this through 
via coordinate transformations and using nearest neighbor relationships to capture object-level details and local contextual information, GFA sub-module uses self-attention mechanisms to selectively aggregate the features of the key voxels across the entire scene for capturing scene-level contextual information. Experiments on nuScenes and KITTI demonstrate the effectiveness of our method. Specifically, on nuScene, SparseDet surpasses the previous best sparse detector VoxelNeXt by 2.2\% mAP with 13.5 FPS, and on KITTI, it surpasses VoxelNeXt by 1.12\% $\mathbf{AP_{3D}}$ on hard level tasks with 17.9 FPS.
\end{abstract}


\IEEEpeerreviewmaketitle

\section{Introduction}

\IEEEPARstart
% 刘林修改: 基于 LiDAR 的 3D 对象检测是自动驾驶中的一项重要任务，随着智能交通系统和自动驾驶技术的发展，得到了广泛的研究。现有的高性能 3D 对象检测器通常在骨干网络和预测头中构建密集特征图 
% 随着雷达传感器数据的可用性导致了Lidar-based 3D目标检测的显著进展，同时因其在自动驾驶 的潜在应用而受到越来越多的关注。目前大多数基于LiDAR的3D检测器利用3D Sparse CNN 提取稀疏体素特征并将其转换为密集的特征图随后进行下一步的预测。然而most methods将稀疏体素特征转换为密集BEV表示，并执行后续的预测。尽管在一些数据集基准上取得了出色的性能，但由于密集检测方案带来的计算成本限制了这些方法在实际中大范围检测场景的应用。因此，研究人员越来越关注稀疏检测框架的设计。
% 目前，针对模型设计思想可将稀疏检测框架为两类，第一类方法遵循中心检测的设计思想~\cite{centerpoint} 将物体中心视为物体的代理，然而由于点云仅分布在物体表面的特性导致了中心点特征缺失的问题~\cite{FSD}，针对上述问题, 这系列方法~\cite{VoxelNeXt,SAFDNet} 旨在利用特征扩散策略将物体表面的特征传递至物体中心处，尽管解决了中心特征缺失的问题，但仅利用单个中心体素作为物体的代理，未聚合邻域信息，导致削弱中心点呈现物体的能力。另一类方法~\cite{FSD,FSDV2}，采用投票机制，将前景点聚集成以对象为中心的集群，以进行进一步的预测。FSD 如何做的， FSDV2如何做的。FSD首先将原始点云分割成前景和背景，然后进行中心投票进行实例聚类。随后，它从每个集群中提取实例特征以进行初始预测，这些特征由组校正头细化。FSDv2消除了聚类任务，并利用对投票点进行虚拟体素划分来进行实例特征提取，但它仍然依赖于点分割和预测细化。尽管具备上下文聚合能力，然而模型过于依赖额外的辅助任务以及大量的超参数，导致推理速度的不佳。
% 
3D object detection is a critical task in autonomous driving, promoting the advances of intelligent transportation systems, and has gained widespread attention ~\cite{wanglisurvey,song2024robustness,graphalign,graphalign++,Multi-sem-fusion,DyFusion,urformer,RI-fusion,robofusion,interfusion,graphbev,Voxelnextfusion,RPFA-Net,wang2022multi,pursuing3D}. With the availability of various sensor modalities, such as cameras and LiDAR, significant progress has been made in single-modal 3D object detection using either camera images~\cite{bevdepth,bevheight,bevheight++,bevdet4d,mixteaching} or LiDAR point clouds~\cite{RA3D,wang2023multi,survery:wangli_mm3dod}. Compared to the image data provided by cameras, LiDAR point clouds offer accurate depth and position information and have led to extensive research %on LiDAR-based detection methods 
in recent years 
%and LiDAR-based methods have achieved better performance than image-based methods 
\cite{pointpillars,SECOND,voxelnet,pointgnn,pointrcnn,vp-net,satgcn,fuzzynms}. 

\begin{figure}[!t]
    \centering
    \includegraphics[width=\linewidth]{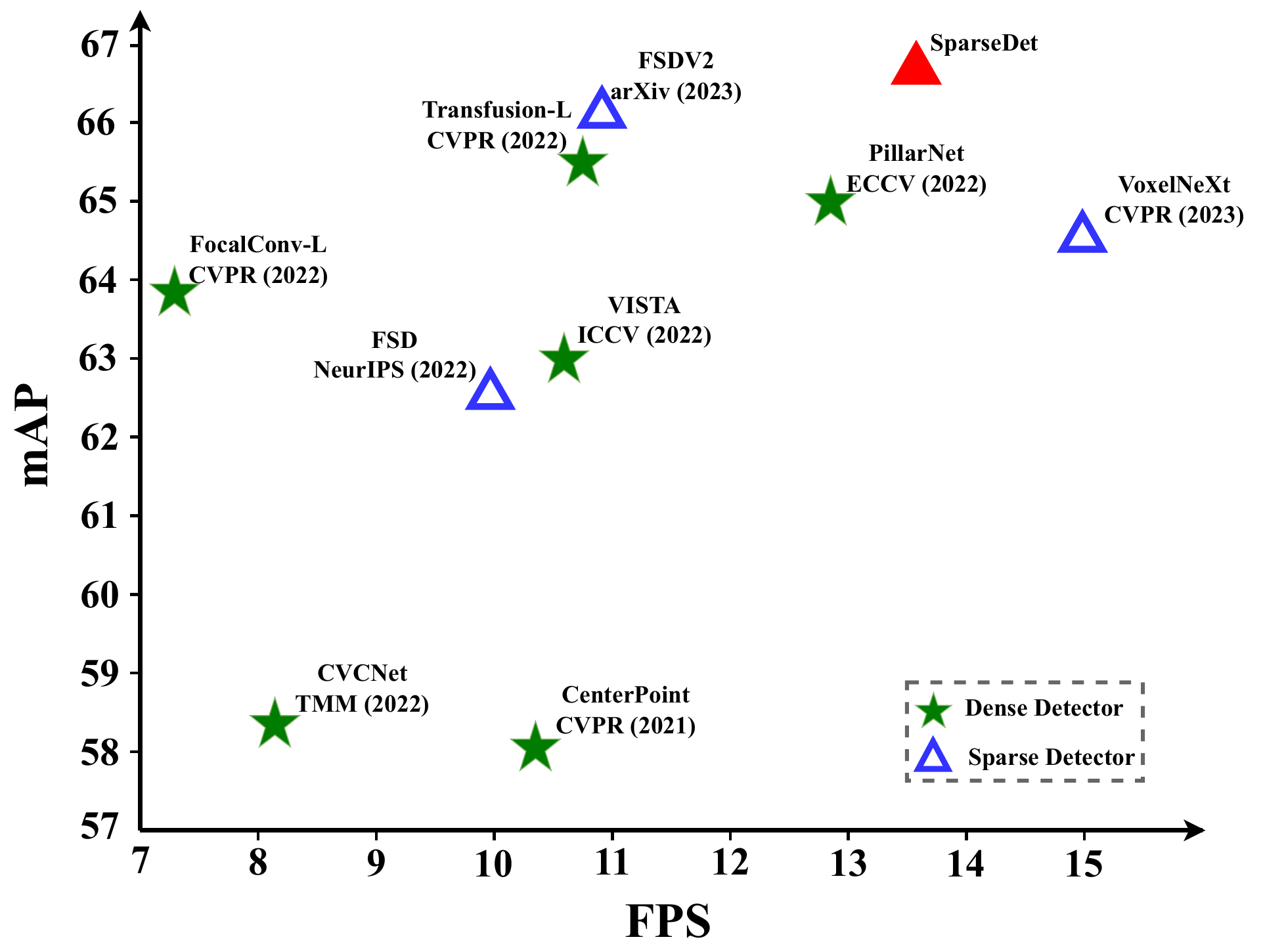}
    \caption{The comparison of SparseDet with existing LiDAR-based detectors on nuScenes\cite{nuscenes} test dataset, where the vertical axis represents mAP, and the horizontal axis represents model inference speed (FPS). Compared to other sparse detectors, SparseDet achieves the highest mAP while maintaining an excellent inference speed.}
    \label{fig:Fps}
\end{figure}

\begin{figure}[!t]
    \centering
    \includegraphics[width=\linewidth]{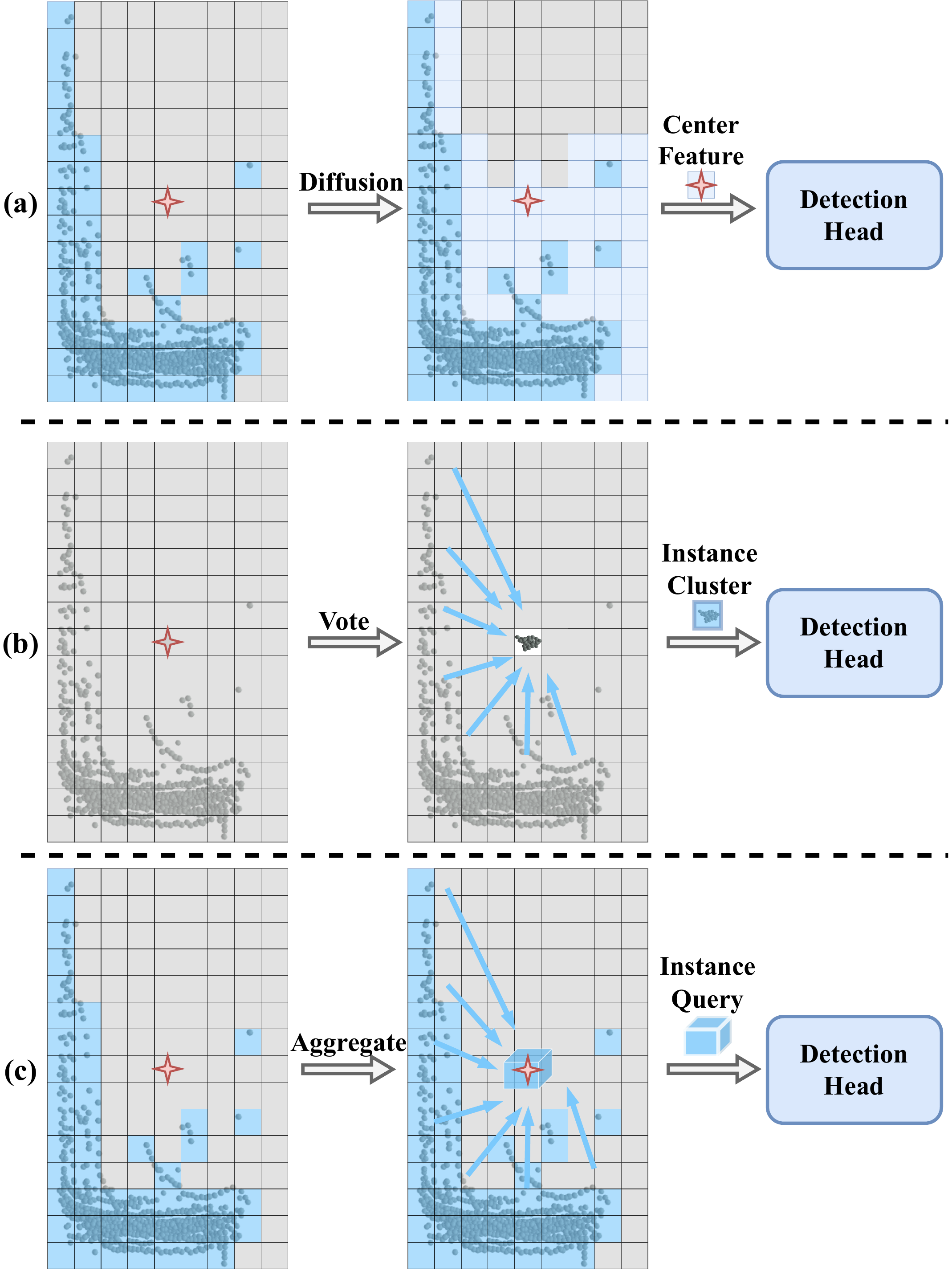}
    \caption{\textcolor{black}{\textbf{Comparison between SparseDet and other sparse detection frameworks~\cite{FSD,FSDV2,voxelnext,SAFDNet}.} (a) %In this series, 
    The first category methods, VoxelNeXt~\cite{voxelnext} and SAFDNet~\cite{SAFDNet}, diffuse feature  by stacking convolutional layers to fill center voxels. However, utilizing only a single central voxel feature as a proxy for an object neglects an amount of instance features, thereby weakening the ability to represent objects based on their central voxels. (b) The second category method, FSD~\cite{FSD} and FSDV2~\cite{FSDV2}, utilize a voting mechanism to aggregate foreground points into object-centered clusters for further prediction. However, these methods overly relies on point segmentation and prediction refinement which results in time delays. (c) Our SparseDet addresses the issue of insufficient information representation in central voxel features by utilizing sparse queries as object proxies and selectively aggregating sparse voxel features at interested positions, avoiding the need for additional auxiliary tasks.}}
    \label{fig:motivation}
\end{figure}

Existing high-performance 3D object detectors~\cite{SECOND,centerpoint,HEDNet,casa} typically leverage 3D sparse CNNs to extract features from sparse voxels, and then convert these features into dense feature maps for detection. Although these methods~\cite{centerpoint,HEDNet,pgd} have demonstrated impressive detection accuracy on a range of benchmark datasets~\cite{nuscenes,kitti,Waymo,dual-radar}, extending them to more practical long-range scenarios becomes challenging. This is because the computational costs associated with the dense feature mappings grow quadratically as the perception range increases. Therefore, some recent works~\cite{FSD,FSDV2,voxelnext,SAFDNet} have attempted to construct fully sparse detectors to cope with this issue. 

Currently, based on the strategy of object proxies, LiDAR-based sparse detectors can be categorized into two distinct classes. The first category (e.g., ~\cite{voxelnext,SAFDNet}) employs stacked convolutional layers to propagate sparse features to central voxels, which are then utilized as object proxies for detection. The second category (e.g.,~\cite{FSD,FSDV2}) aggregates foreground points into clusters and treats them as object proxies. Although the aforementioned methods have successfully constructed fully sparse detectors, the first category only treats individual central voxels as object proxies, lacking the ability to learn point cloud context. This weakens the information representation capacity of object proxies. As shown in Fig.~\ref{fig:motivation} (a), treating only the central voxel features as object proxies leads to the loss of some point cloud information from the same instance. The second category clusters foreground points into object-centric clusters for further prediction. As shown in Fig.~\ref{fig:motivation} (b), it (e.g., ~\cite{FSD} or ~\cite{FSDV2}) initially segments the raw point cloud into foreground and background regions, performs center voting for instance clustering, then extracts instance features from each cluster for initial predictions. %which are refined by a grouping refinement head.
Although these methods possess contextual aggregation capabilities, they heavily rely on additional auxiliary tasks and numerous hyperparameters, resulting in poor inference speed.%FSDV2~\cite{FSDV2} eliminates the clustering task and utilizes virtual voxel partitioning to extract instance features from the voting points. 

In order to achieve efficient detection while effectively aggregating contextual information in sparse frameworks, in this study,  we propose a simple and effective fully sparse 3D object detection framework called SparseDet. SparseDet utilizes a 3D sparse convolutional network to extract features from point clouds and transforms them into 2D sparse features for further prediction via detectio
n head. As shown in Fig.~\ref{fig:motivation} (c), SparseDet designs sparse queries as object proxies, allowing to flexibly and selectively aggregate point clouds to obtain object proxies in a scene. Compared to the previous sparse aggregation paradigm~\cite{FSD,FSDV2}, firstly, SparseDet extends the aggregation of local contextual information to multi-scales feature space, thereby obtaining richer local information. Furthermore, in contrast to prior methods~\cite{FSD,FSDV2} that only focus on aggregating foreground point features, SparseDet can aggregate the scene-level context for each instance to facilitate potential collaboration between the scene and instance features. Finally, SparseDet does not require any additional auxiliary task.

In order to enhance SparseDet's ability to learn point cloud context, we have designed two key modules, the Local Multi-scale Feature Aggregation (LMFA) module and the Global Feature Aggregation (GFA) module. The LMFA module is aimed to capture local contextual information at multiple scales by only leveraging simple coordinate transformations and voxel's nearest-neighbor relationships to collect features from neighboring key voxels. These features are then aggregated to obtain richer local representations. By these operations, LMFA module enables SparseDet to capture fine-grained details and local variations in point clouds. This enhances the expressive power of sparse queries and enables better representation of the underlying structures within point clouds. Afterwards, we initialize the aggregated features of the key voxels as sparse queries and feed them into GFA module. GFA module targets to aggregate global point cloud information across the entire scene by utilizing global sparse queries in a larger receptive field. % including a wider context.  
This allows SparseDet to have a comprehensive understanding of the scene and incorporates global context into the detection process. Benefiting from the comprehensive learning of contextual information, SparseDet significantly enhances the information representation capability of object proxies and demonstrates superior performance (please see Fig.~\ref{fig:Fps} as a reference).

% 我们在KITTI和Nus两个数据集验证了我们方法的有效性。相比大多数SOTA方法基，SparseDet展现出了更优异的性能。NUs数据集的远程目标检测上，SAFDNet比Voxelnext高出2.6% mAP，同时不增加显著的延迟，在性能优于FSDV2的同时，模型推理速度快1.38倍。这些结果证明了SparseDet在需要远程检测的情况下的有效性。

We conduct extensive experiments on two popular datasets KITTI~\cite{kitti} and nuScenes~\cite{nuscenes} to verify the effectiveness of our method. Compared to the most of state-of-the-art (SOTA) methods, SparseDet demonstrates superior performance. Specifically, on the large-scale nuScenes dataset, SparseDet achieves a 2.2\% improvement in mAP compared to the baseline model VoxelNeXt. And on KITTI SparseDet, it surpasses VoxelNeXt by 1.12\% $AP_{3D}$ on hard level tasks. Significantly, this performance enhancement is accomplished without introducing substantial additional latency. While achieving better performance than FSDV2, 
SparseDet attains an FPS of 13.5, which is 1.38 times faster than FSDV2. 

% 最后一段修改前 4.28下午张国欣改
% To address the issues associated with existing voxel-based multi-modal 3D object detection, 
% % particularly regarding projection schemes, 
% we propose a simple, unified, and effective multi-modal fusion framework, dubbed~\textbf{\textcolor{black}{VoxelNextFusion}}. 
% First, we follow the principle of effective fusion at both coarse and fine levels by using a simple multi-layer perceptron (MLP) to obtain fused features with both image continuity and semantics from one-to-one and one-to-many projections, as shown in Fig.~\ref{fig_second_case}. 
% Second, we differentiate between foreground and background features to eliminate any potential impact of background pixel features, which further improves the exploitation of important features in the fusion process. 
% Finally, our proposed strategy significantly improves the fusion performance in existing voxel-based methods, particularly when the object contains only a few points. 
% We outperform previous state-of-the-art methods with KITTI~\cite{kitti} and nuScenes~\cite{nuscenes}, particularly on long-range objects. 
% 我们sparsedet的框架，首先，我们将点云体素化并放入3d稀疏卷积骨干网中。然后，我们对骨干网中最后三层的稀疏体素特征执行高度压缩~\cite{voxelnext}得到F4,F5,F6。In LMFA，我们concat F4,F5,F6 得到 F7 并进行关键体素位置的预测，随后，根据坐标转换关系，我们将关键体素特征转换到F4,F5,F6的空间中并根据最近邻关系聚合邻域体素特征。随后我们将聚合好的体素特征根据索引放回F7中。In GFA module, SparseDet将稀疏体素作为query,自适应地全局聚合稀疏体素特征，同时, 尺度自适应权重图使query更关注关键位置附近的特征。 随后聚合好的query被送入FFN进行结果的预测。 Adaption fusion 代表自适应融合 as shown in Fig. 
\begin{figure*}[!t]
    \centering
    \includegraphics[width=1\linewidth]{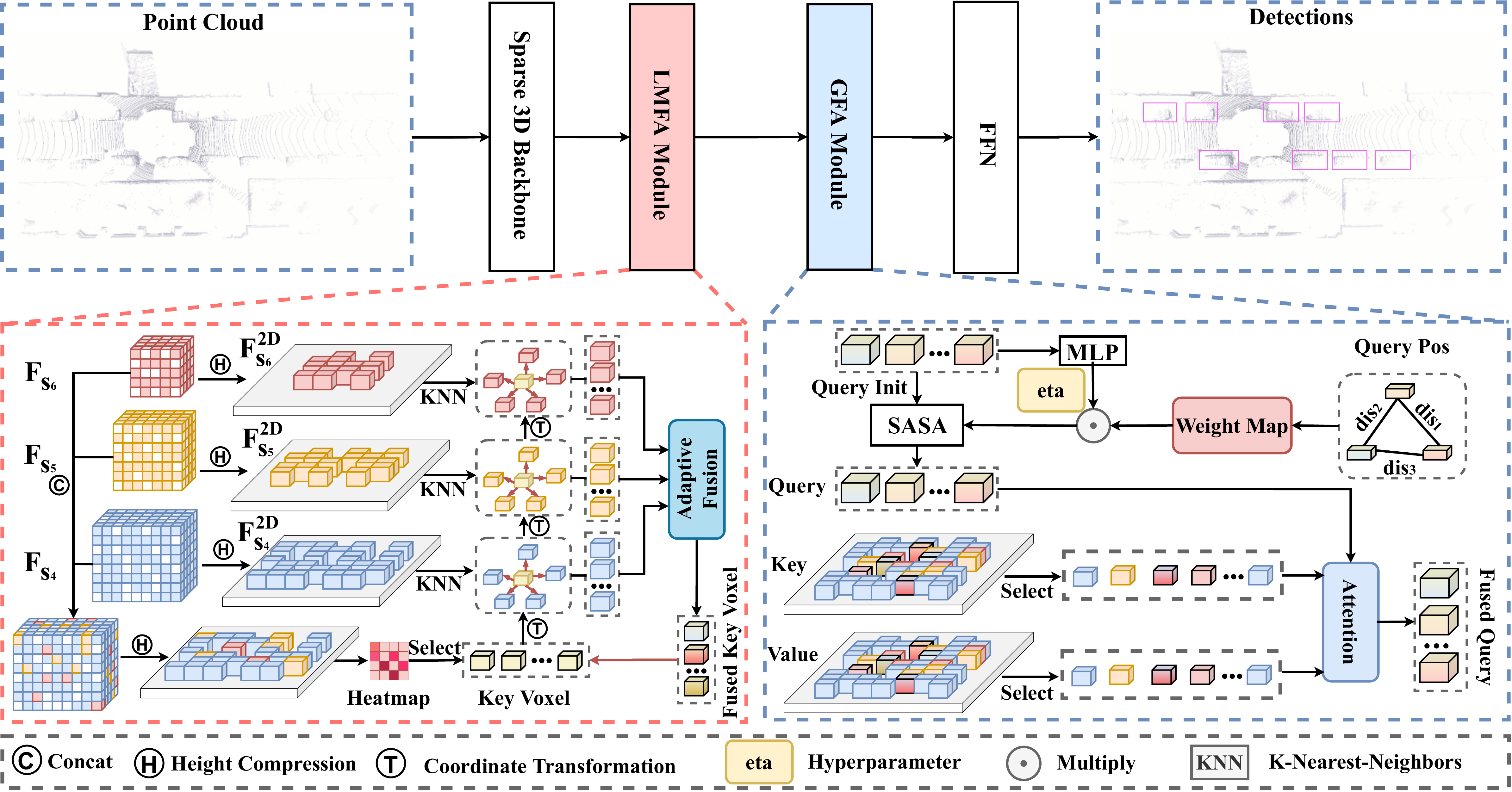}
    \caption{The framework of our \textcolor{black}{SparseDet}. First, we voxelizes point clouds and feed voxels into a 3D sparse convolution backbone. Then, we perform high compression~\cite{voxelnext} on sparse voxel features ($F_{s_{4}}$, $F_{s_{5}}$, $F_{s_{6}}$) of the last three layers in the 3D backbone to obtain 2D sparse features of these three layers, denoted as $F_{s_{4}}^{2D}$, $F_{s_{5}}^{2D}$ and $F_{s_{6}}^{2D}$. In LMFA, we concat $F_{s_{4}}$, $F_{s_{5}}$ and $F_{s_{6}}$ to obtain $F_{Fusion}$. After applying high compression~\cite{voxelnext} on $F_{Fusion}$, we have $F_{Fusion}^{2D}$ and then perform key voxel position prediction using a heatmap. Through coordinate transformation, we convert the key voxel features to the spaces of $F_{s_{4}}$, $F_{s_{5}}$ and $F_{s_{6}}$ and aggregate the neighborhood voxel context based on $K$ Nearest Neighbor (KNN) relationships. Subsequently, we replace the aggregated voxel features back into $F_{Fusion}$ based on the indices to enhance the feature representation capability of sparse features. In GFA module, we utilizes sparse voxels as queries to adaptively aggregate global sparse voxel features, where scale-adaptive weight map allows queries to autonomously learn the receptive field for aggregating information from relevant positions. At last, the aggregated queries are fed into FFN for result prediction. Adaption Fusion means the adaptive fusion of multi-scale features and FFN is a Feed Forward Neural Network.}
    \label{fig:framework}
\end{figure*}

\section{Related work}
% 这一段要引出稀疏检测的优势
\subsection{LiDAR-based Dense Detectors}
% 尽管点云的稀疏特性与2D数据不同，但在设计3D探测器时，常常采用与2D探测器相似的方法，许多研究都使用二维密集检测头来解决3D检测问题。
%  voxelnet将点云划分为规则的网格并利用密集的3D 骨干网进行特征提取，随后 applies dense region proposal network and head for prediction. 随后的SECOND通过构建哈希表实现了稀疏卷积[417]和子流形卷积[418]算子的高效计算来加速VoxelNet,但仍需要密集的BEV特征图以及密集检测头来完成检测任务。随后的网络大多遵循 Second 3D sparse backbone + 2D Dense detection head 这一范式。
% 尽管LiDAR Dense Detectors在多个数据集基准上表现出了出色的性能。然而对于密集BEV特征图以及密集检测头的依赖使其难以扩展到远距离的场景检测中。这是因为密集BEV特征图的计算成本随检测距离而二次增加。这极大的限制了LiDAR Dense Detectors在现实场景的应用。
Although point cloud data exhibits different sparsity characteristics compared to 2D image data, 3D object detectors are often designed by referencing 2D detectors. For example, most works~\cite{voxelnet,centerpoint,SECOND,voxelfpn,voxelrcnn,hannet} have utilized 2D dense detection heads to address the problem of 3D detection.  These methods are usually called LiDAR-based dense detectors~\cite{SAFDNet}.

As a pioneer, VoxelNet~\cite{voxelnet} partitions point clouds into regular grids and utilizes a 3D backbone network for feature extraction. It then applies a dense head for prediction. Based on VoxelNet, SECOND~\cite{SECOND} implements efficient computation of sparse convolution and submanifold convolution operators to gain fast inference speed by constructing a hash table. However, SECOND still requires dense Bird's Eye View (BEV) feature maps and dense detection head for detection. On the influence of SECOND, most subsequent networks~\cite{voxelrcnn,centerpoint,voxelfpn,pvrcnn,parta2,afdetv2,focalconv} follow the paradigm of utilizing a 3D sparse backbone combined with a 2D dense detection head.

Although LiDAR-based dense detectors have shown excellent performance on multiple benchmark datasets~\cite{kitti,nuscenes,Waymo,dual-radar}, their reliance on dense Bird's Eye View (BEV) feature maps and dense detection heads makes them be challenging to scale-up to long-range scene detection. This is because the computational costs of dense BEV feature map increases quadratically with detection distance~\cite{FSD}. This drawback significantly restricts the practical applications of LiDAR-based dense detectors in real-world scenarios.
\subsection{LiDAR-based Sparse Detectors}
% Point-based 方法通过在点云中选择有意义的关键点进行特证聚合以及检测,不需要对整个空间进行密集的采样和计算，故是天然的全稀疏检测器。 FSD和FSDV2是这一系列方法的代表。FSD通过对分割的前景点进行聚类来表示单个对象。然后，它将PointNet提取的特征输入到检测头中进行校准和预测。 尽管充分聚合了前景信息，然而由于依赖额外的辅助任务以及大量的超参数，导致推理速度的不佳

Currently, sparse detectors include point-based methods and partial voxel-based methods. Point-based methods~\cite{pointgnn,pointpainting,pointrcnn,fastpointrcnn} utilize key points within point clouds for feature aggregation and detection. These methods do not require dense sampling and computation across the entire space, making them inherently sparse detectors. FSD~\cite{FSD} and FSDV2~\cite{FSDV2} are representative of this series of methods. FSD represents individual objects by clustering segmented foreground points. It then fed features which are extracted by PointNet into a detection head for calibration and prediction. In FSDv2, the instance clustering step is replaced with a virtual voxelization module, which aims to remove the inherent bias introduced by manually crafted instance-level representations. Despite adequately aggregating foreground information, the reliance on additional auxiliary tasks and numerous hyperparameters leads to poor inference speed.

% voxelnext 为模型引入额外的下采样层来将体素放置到物体中心附近，并随后对关键体素进行特征扩散来将特征传递到物体中心处。 SAFDNet在VoxelNext的基础上提出了自适应特征扩散策略来解决中心特征缺失问题。尽管取得了优异的效率，然而由于缺乏聚合邻域特征的能力，仅依赖单个中心体素特征进行检测极大的削弱了物体代理的信息表达能力导致模型的性能下降。

Among the voxel-based sparse methods, VoxelNeXt~\cite{voxelnext} introduces additional downsampling layers to place voxels near the centers of objects and subsequently performs feature diffusion on key voxels to propagate features towards the object centers. SAFDNet~\cite{SAFDNet} addresses the issue of missing central features by proposing an adaptive feature diffusion strategy. Although SAFDNet~\cite{SAFDNet} and VoxelNeXt~\cite{voxelnext} have achieved impressive efficiency, they solely rely on single center voxel features for detection, which significantly weakens the information representation capability of object proxies, ultimately leading to a decline in model performance. As mentioned before, treating only the central voxel features as object proxies leads to the loss of some point cloud information from the same instance as illustrated  in Fig.~\ref{fig:motivation} (a). In this work, we use sparse queries and attention mechanism to obtain object proxies by LMFA and GFA modules, which enable to dynamically capture contextual information at different granularities. This promotes the collaboration between scene-level and instance-level features, thereby enabling the model to obtain richer and more accurate object representations. %In this work, we use sparse queries and attention mechanism to obtain object proxies, and leverage LMFA and GFA modules to dynamically capture contextual information at different granularities and aggregate it into object detection proxies, promoting the collaboration between scene and instance features, thereby enabling the model to obtain richer and more accurate target representations.

% 总结下我们方法的优势

\section{\textcolor{black}{SparseDet}}
\label{sec:method}
% 在本章中，我们提出了一个简单efficientive的 端到端的sparse 检测框架。通过自适应充分聚合sparse 体素的邻域information , 旨在解决sparse 检测框架 center feature 特征表达能力不足的问题。Fig.中展示了我们sparseDet的结构。 为了充分聚合 sparse 物体代理的context in SparseDet, 我们设计量两个 sub-modules namely LMFA modules(Local Multi-scale Feature Aggregation) and GFA modules( Global Feature Aggregation). 
%In this section, we propose a simple and efficient LiDAR-based sparse detection framework. By adaptively and comprehensively aggregating neighborhood information of sparse voxels, our framework aims to address the issue of insufficient feature representation capability in the center feature of the sparse detection framework. Fig.~\ref{fig:framework} illustrates the structure of our SparseDet. To fully aggregate the contextual information of sparse object proxies, we design two sub-modules, namely LMFA (Local Multi-scale Feature Aggregation) modules and GFA (Global Feature Aggregation) modules.

In this section, we propose a simple and efficient LiDAR-based sparse detection framework SparseDet.  Fig.~\ref{fig:framework} illustrates its structure which follows the pipelines of the fully sparse network VoxelNeXt~\cite{voxelnext}. But differently, in order to fully aggregate the contextual information in point clouds to enhance the information exression capability of the sparse object proxies, we design two sub-modules, LMFA (Local Multi-scale Feature Aggregation) module and GFA (Global Feature Aggregation) module. The two modules intend to adaptively aggregate multi-level contextual information on point clouds, and make SparseDet be able to strongly enhance the information representation capability of object proxies so as to improve the performance of 3D detection with low computational costs.

\subsection{Local Multi-scale Feature Aggregation}
\begin{figure}[!t]
    \centering
    \includegraphics[width=1\linewidth]{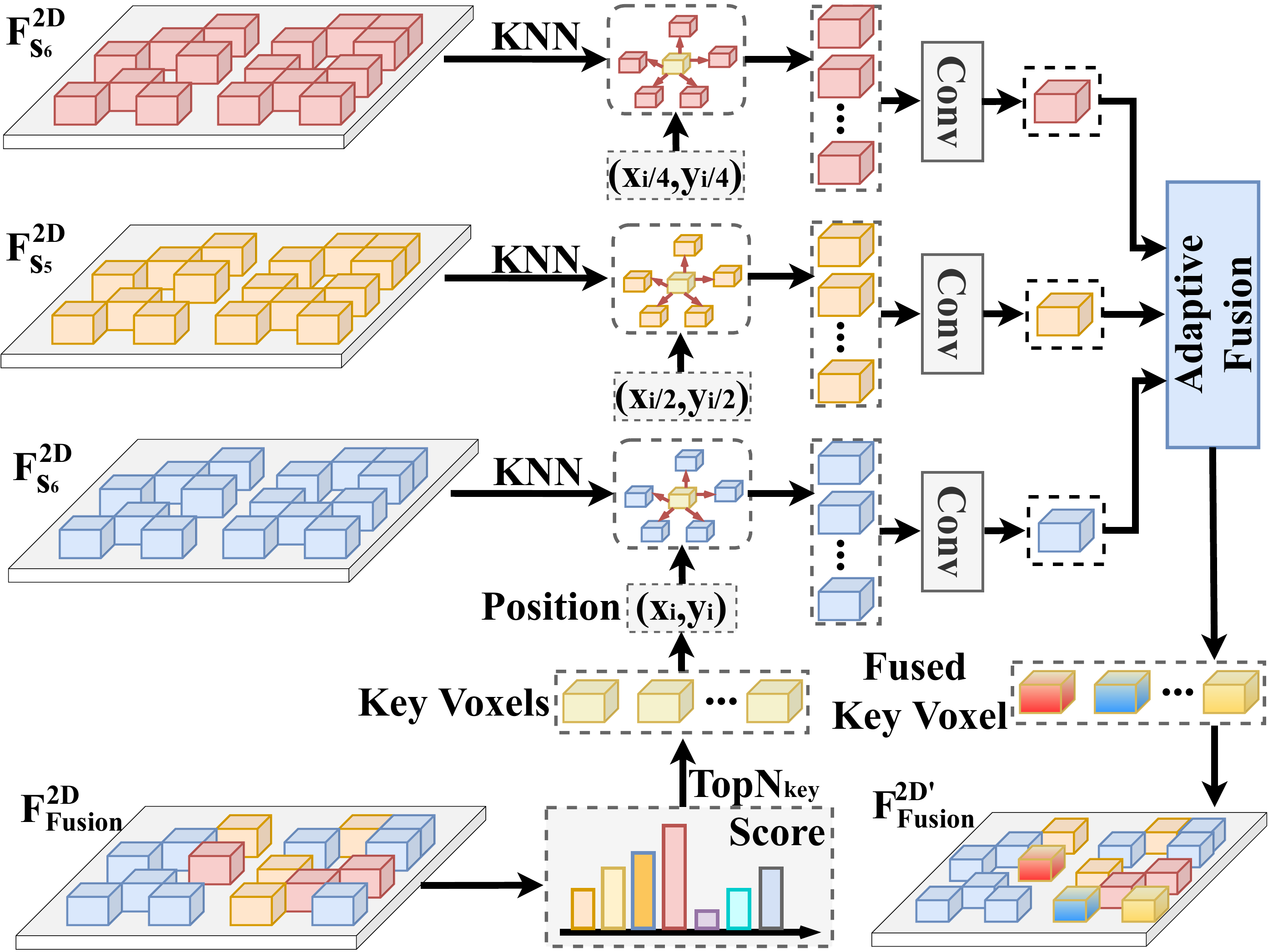}
    \caption{The architecture details of \textbf{LMFA}.The LMFA module performs heatmap prediction from the $F_{Fusion}^{2D}$, and selects the top-$N_{key}$ high-scoring voxels as key voxels. It then converts the position information of the key voxels to their original feature spaces according to the downsampling ratios. Based on the position information of the key voxels, LMFA module finds the KNN voxel features and fuses them by using $Conv_{1*1}$. Finally, the neighborhood features from multi-scale feature spaces are fed into the adaptive-fusion module for adaptive fusion.}
    \label{fig:localfusion}
\end{figure}

% 现有的Lidar-based sparse 检测方案大多利用center 体素特征作为物体代理并to do 检测。尽管将center特征视为物体代理提供了精确的位置information,然而单个 center 体素特征不足以充分涵盖整个物体的information. 如图Fig 1 a 所示,大量的物体information 无法聚合到center 体素位置. 这严重损害了物体代理的表达能力。因此，我们提出了 LMFA module 来弥补这一个缺点。如图localfusion 所示, 通过最近邻位置 relation 我们动态地聚合关键体素邻域information来增强其特征的表达能力. 值得注意的是，注意到3D空间物体尺度的分布差异，我们将LMFA扩展至多尺度空间中。LMFA主要由三个部分组成: Sparse key voxel candidate selection, Fusion of voxel features from different scales. 
%Existing voxel-based multi-modal methods typically use a one-to-one mapping between voxels and images for fusion. While the camera pixel that uniquely corresponds to each voxel can be precisely located, LiDAR features represent a subset of points contained within a voxel, so their corresponding camera pixels lie within a polygon. The one-to-one mapping loses the original intention of using images, namely semantic and continuous properties, which is even worse for long-range detection. Therefore, we propose P$^2$-Fusion (Patch-Point Fusion) to compensate for the shortcomings. As shown in Fig.~\ref{fig:framework}, after voxelization of the original point cloud, multiple layers of 3D sparse convolution encoding are performed. We implement our proposed P$^2$-Fusion between the first and second layer encodings. P$^2$-Fusion is primarily composed of two stages: \textbf{Projection}, and \textbf{Fusion}.
Most LiDAR-based sparse detection methods~\cite{voxelnext,FSD,FSDV2,SAFDNet} utilize center voxel features as object proxies for detection. Although using center features as object proxies provides accurate positional information, a single center voxel-feature is insufficient to fully capture the entire information of an object. %as shown in Fig.~\ref{fig:motivation} (a).
This severely weakens the expressive capability of the object proxies. Therefore, we propose LMFA module to compensate for the shortcomings. In the LMFA module, we focus on learning the local contextual information around an object, which helps to understand details such as the shape, size, and relative position of a target object. As shown in Fig.~\ref{fig:localfusion}, we dynamically aggregate neighborhood information of key voxels through $K$ nearest neighbor (KNN) positional relations to enhance their feature representation capability. The aggregated key voxel features will then be used to initialize the sparse object queries. It is worth noting that considering the distribution differences in 3D object scales, we extend LMFA to multiple-scale spaces. Thus, LMFA primarily consists of two steps, \textbf{Sparse Key Voxel Selection} and \textbf{Fusion Of Voxel Features From Different Scales}.
\subsubsection{Sparse Key Voxel Selection}
% 首先，我们将点云体素化并放入3d稀疏卷积骨干网中. follow voxelnext 我们为Sparse CNN backbone network~\cite{second}中额外添加了两个下采样层，这一步骤具备两个关键作用: 一方面 通过额外的downsampling过程构建了多个特征尺度空间以便一步的LMFA的特证聚合，其次，通过额外的采样过程以及after地高度压缩操作 我们可以为物体center空白处放置体素特征以更准确地构建邻域relation。by 上述操作，这样原始的Sparse CNN 由4阶段 {S1，S2, S3, S4} 变为 {S1,S2,S3,S4, S5, S6} with the feature strides {1, 2, 4, 8, 16, 32}。每一阶段的3D 稀疏体素特征可以表示为。After that ，我们将F5,F6体素特征转换到F4特征空间中并将f4,f5,f6 特征concat到一起得到F_fusion，为了消除之前点云数据增强导致的特征错位的影响，体素通过逆运算转换为其原始坐标，例如去除翻转和裁剪。 然后我们对f_fusion, f4, f5, f6 执行高度压缩得到 f_fusion, f4, f5, f6。具体地，我们follow voxelnext 将所有体素放在地面上，并在相同的位置求和。 % 具体地流程如下列伪代码所示。
First, we voxelize point clouds and feed them into a 3D sparse convolutional backbone network. Referring to VoxelNeXt~\cite{voxelnext}, we add two additional downsampling layers to the 3D sparse backbone network~\cite{SECOND}. This step serves two key purposes. Firstly, it constructs multi-scale feature spaces through the additional downsampling process to facilitate the subsequent feature aggregation in LMFA module. Secondly, through the additional sampling and height compression operations, we can place voxel features in object centers which are blank to construct neighborhood relations more accurately. With the above operations, the original Sparse 3D convolutional backbone transitions from $\left \{ F_{s_{1}},F_{s_{2}},F_{s_{3}},F_{s_{4}}  \right \} $ to $\left \{ F_{s_{1}},F_{s_{2}},F_{s_{3}},F_{s_{4}},F_{s_{5}},F_{s_{6}}  \right \} $ with feature strides $\left \{ 1,2,4,8,16,32 \right \} $, where $F_{s_{i}}\in R^{N_{i}\times C_{i}}$ represents the 3D sparse voxel features for each stage, $s_{i}$ means the $i$-th stage, $N_{i}$ is the number of non-empty voxels, and $C_{i}$ is the number of channels.  Afterward, we transform $F_{s_{5}}$ and $F_{s_{6}}$ to the feature space of $F_{s_{4}}$ and concatenate $F_{s_{4}}$, $F_{s_{5}}$ and $F_{s_{6}}$ together to obtain $F_{Fusion}$. %to eliminate the impact of feature misalignment caused by previous data augmentation. %$F_{Fusion}$ is transformed back to their original coordinates through an inverse operation, such as removing cropping and flipping~\cite{pointaugmenting}. 
Then, we perform high compression on $F_{Fusion}$, $F_{s_{4}}$, $F_{s_{5}}$ and $F_{s_{6}}$ to obtain $F_{Fusion}^{2D}$, $F_{s_{4}}^{2D}$, $F_{s_{5}}^{2D}$ and $F_{s_{6}}^{2D}$. Specifically, following the VoxelNeXt~\cite{voxelnext}, we replace all the voxel features on the ground plane and sum them up at the same positions. %The workflow is demonstrated in Alg.~\ref{algorithm:MSF}.
% 具体的流程如算法MSF所示
To select key voxels, we use heatmap operation which predict voxel scores $Score\in R^{N_{voxel} \times Cls}$ for $Cls$ classes based on the sparse voxel feature $F_{Fusion}^{2D}$, where $N_{voxel}$ respresents the number of non-empty voxels in $F_{Fusion}^{2D}$. We designate the voxels closest to an object center as positive samples and utilize FocalLoss~\cite{centerpoint} for supervision. This means that voxels with higher scores have a higher probability of belonging to the foreground. Subsequently, we apply the top-$N_{key}$ score operation to $F_{Fusion}^{2D}$ to obtain $N_{key}$ key sparse voxel candidates. Here, $N_{key}$ is set to a default value of 500.

\subsubsection{Fusion of Voxel Features From Different Scales}
% 在本节中，我们通过构建最近邻图来获得稀疏候选体素在不同尺度的邻域信息以实现更充分的上下文建模，旨在捕捉体素之间的局部关系来提供更丰富的特征表示从而弥补少量稀疏特征表征能力不足的问题。

% After Sparse Voxel Candidate Selection, 我们得到了N个关键稀疏体素的特征，定义为F_{key} N*C，其中C为特征的通道数。And 对应的坐标位置索引定义为I_{key} = (N,2) ，where 2代表2D位置索引。我们首先将关键体素的在S4尺度空间的位置坐标I_{s_{4}}分别除以2以及4使其转换到更低分辨率的{S5,S6}体素空间中并保存对应的空间坐标索引{I_{s_{5}},I_{s_{6}}}。随后给定N个稀疏体素在不同尺度空间的位置坐标信息，我们为其寻找Km个距离最近的稀疏体素，其中km的取值随尺度空间的变化而减半(加一个公式)。Km默认取16。值得注意的是，为进一步节约计算资源，我们将稀疏体素的空间位置按照KD树的数据结构组织以实现高效的query. 

% 给定编码好的关键稀疏体素的多尺度特征，一种朴素的融合方案是连接多尺度特征以形成单个特征。然而我们注意到一些物体的检测更多依赖来自特定尺度的信息，但并非总是来自所有尺度的信息。As shown in Tab 1. 当缺乏低分辨率的特征信息使，行人类别的指标并未受到过多影响。与传统的FPN不同，我们建议使用可学习的尺度注意力为每个框自动选择尺度。每个对象查询生成尺度选择的注意力权重为，其中O为候选体素特征。最终稀疏体素候选特征具体由尺度选择权重α16、α32、α64自适应加权和聚合
\ %注意此处还有一空格
%\newline %新起一行
% \indent 
In this section, we construct a $K$ nearest neighbor graph to obtain neighborhood information for sparse candidate voxels at different scales to gain more comprehensive local contexts which address the insufficient information representation capacity of sparse features. 

After sparse key voxel selection, we obtain the features of $N_{key}$ sparse voxels, denoted as $F_{key}\in R^{N_{key} \times C}$, where $C$ represents the number of channels. The corresponding coordinate position indices are defined as $I_{key}$, with a shape of ($N_{key}, 2$), which represents the 2D position indices. We first divide the position coordinates of the $N_{key}$ voxels in the $S_{4}$ scale, denoted as $I_{s_{4}}$, by 2 and 4, respectively, to transform them into the lower-resolution voxel spaces of $\left \{ S_{5},S_{6} \right \}$. We then save the corresponding spatial coordinate indices as $I_{s_{5}}$,$I_{s_{6}}$. Given the position coordinate information of $N_{key}$ sparse voxels in different scale spaces, we aim to find the $K$ nearest voxels $N_{Neighbor}^{s_{i}}$  for each key voxel. The value of $N_{Neighbor}^{s_{i}}$ is halved as the scale space changes, which can be determined using the following formula.
\begin{equation}
\label{eq:voxelset}
\begin{aligned}
    N_{Neighbor}^{s_{i}} = M/2^{i-4}, i = 4,5,6, \\
\end{aligned}
\end{equation}
where the parameter $M$ is set to 8 by default. 

To improve the efficiency of LMFA, we employ the \textit{KD-Tree} algorithm to obtain the indices of $N_{Neighbor}^{s_{i}}$ neighbors for each key voxel at a specific scale $S_{i}$, which are denoted as $I_{Neighbor}^{s_{i}} \in R^{N_{key}\times N_{Neighbor}^{s_{i}} \times 2}$. The surrounding neighbor voxel features $F_{Neighbor}^{s_{i}} \in R^{N_{key}\times N_{Neighbor}^{s_{i}} \times C}$ are obtained by indexing $F_{s_{i}}^{2D}$ with $I_{Neighbor}^{s_{i}}$. Then, MLP is utilized to aggregate the features of the neighbor voxel features $F_{Neighbor}^{s_{i}}$, which is performed by the following formula.
\begin{equation}\label{eq:voxel_agg}
    \begin{aligned}
        F_{Neighbor'}^{s_{i}} = MLP(F_{Neighbor}^{s_{i}}), \\
    \end{aligned}
\end{equation}
where $F_{Neighbor'}^{s_{i}} \in R^{N_{key}\times 1\times C}$ represents aggregated features. By applying the aforementioned feature aggregation scheme at each feature scale in $\left \{ S_{4},S_{5},S_{6} \right \}$, we can obtain aggregated features of multiple scales $F_{Neighbor} \in R^{N_{key}\times N_{scale} \times C}$, where $N_{scale}$ is the number of scales. 

Given the encoded multi-scale features of sparse voxels, a naive fusion approach is to concatenate the multi-scale features to form a single feature~\cite{voxelfpn}. However, we observe that some object detections rely more on information from specific scales rather than from all scales. For instance, low-resolution feature maps lack information about small objects. Therefore, key voxels related to small objects should more effectively gather information solely from high-resolution feature maps. We propose utilizing learnable scale weights to automatically select the scale for each key voxel $F_{key}$ as follows. 
\begin{equation}\label{eq:weight_generate}
    \begin{aligned}
        W_{1}, W_{2}, W_{3} = Softmax(FC(F_{key})),
    \end{aligned}
\end{equation}
\begin{equation}\label{eq:key_voxel_fusion}
    \begin{aligned}
        F_{key} = Conv\left (Concat\left (  F_{key}, \sum_{i=1}^{N_{scale}} W_{i}F_{neighbor'}^{s_{i}} \right ) \right ).
    \end{aligned}
\end{equation}
Where $W_{1}$, $W_{2}$, $W_{3}$ stand for the importance of selecting $F_{neighbor'}^{s_{4}}$, $F_{neighbor'}^{s_{5}}$, $F_{neighbor'}^{s_{6}}$, $F_{key}$ is adaptively aggregated according to the scaled attention weights $W_{i}$ obtained in Eq. (\ref{eq:weight_generate}). These scale weights can be learned during the training process and enable adaptive scale selection based on the characteristics of individual key voxels. Thus, we also call the step adaptive fusion.

% 为了提高模型的处理效率，我们使用KD-Tree算法来获得每个关键体素在特定尺度S_{i}上k个邻居的索引，定义为I. 然后，我们通过索引周围邻域深度来获得多个模型的特征。 随后，我们利用MLP将多个邻域体素特征进行聚合 by the following formula. 通过在每个特征尺度执行上述特征聚合方案。我们可以得到来自多个尺度的聚合特征。例如，低分辨率特征图 E64 中缺少小物体的信息。因此，负责小物体的对象查询应该更有效地仅从高分辨率特征图中获取信息。我们建议利用可学习的尺度权重来自动选择每个关键体素的尺度。这些尺度权重可以在训练过程中学习，并能够基于单个关键体素的特性进行自适应尺度选择。每个对象查询生成尺度选择的注意力权重为，其中O为候选体素特征。最终稀疏体素候选特征具体由尺度选择权重α16、α32、α64自适应加权和聚合。Ci,j根据Eq.(7)中得到的缩放注意权重αj对Ci,j进行加权聚合。通过这种尺度选择注意机制，软选择与每个对象查询最相关的尺度，同时抑制来自其他尺度的视觉特征

% where M默认设置为8，After Sparse Voxel Candidate Selection, 我们得到了N个关键稀疏体素的特征，定义为F_{key} N*C，其中C为特征的通道数。And 对应的坐标位置索引定义为I_{key} = (N,2) ，where 2代表2D位置索引。我们首先将关键体素的在S4尺度空间的位置坐标I_{s_{4}}分别除以2以及4使其转换到更低分辨率的{S5,S6}体素空间中并保存对应的空间坐标索引{I_{s_{5}},I_{s_{6}}}。随后给定N个稀疏体素在不同尺度空间的位置坐标信息，我们为其寻找Km个距离最近的稀疏体素，其中km的取值随尺度空间的变化而减半(加一个公式)。Km默认取16。值得注意的是，为进一步节约计算资源，我们将稀疏体素的空间位置按照KD树的数据结构组织以实现高效的query

\begin{figure}[!t]
    \centering
    \includegraphics[width=1\linewidth]{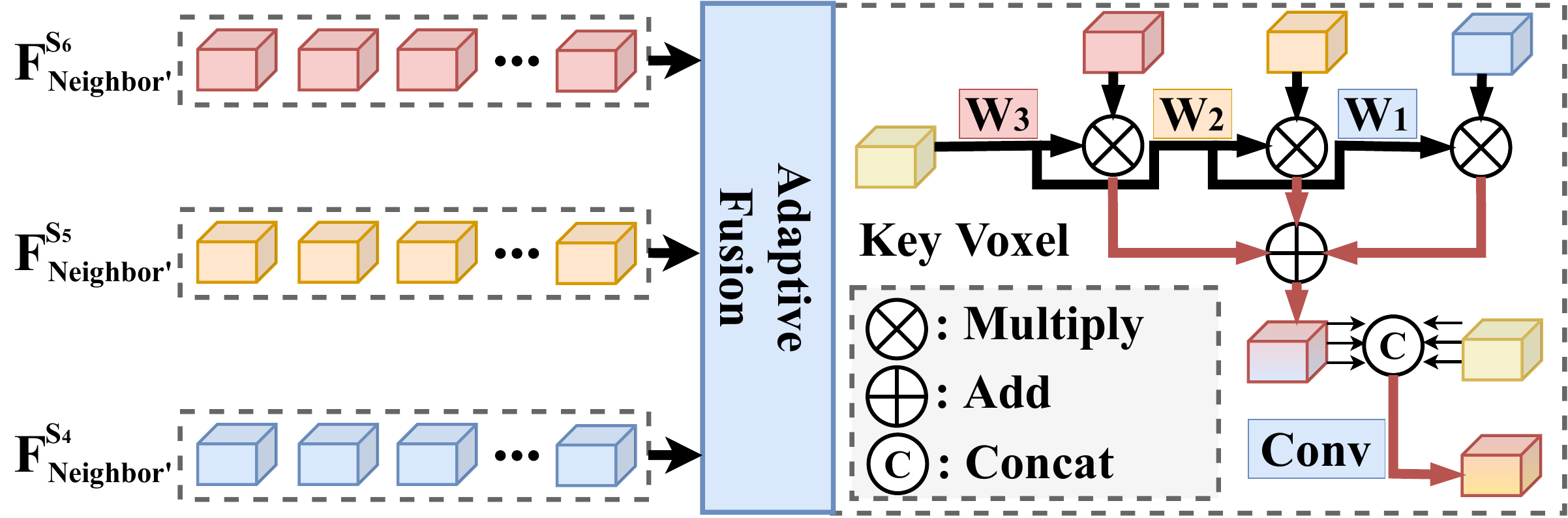}
    \caption{The architecture of \textbf{Adaptive Fusion}, which adaptively weights and fuses neighborhood features ($F_{Neighbor^{'}}^{s_{4}}$, $F_{Neighbor^{'}}^{s_{5}}$ and $F_{Neighbor^{'}}^{s_{6}}$) from multi-scale feature spaces for each key voxel.}
    \label{fig:adaptivefusion}
\end{figure}

With such a scale-selection mechanism, the scale most related to each key voxel is softly selected while the visual features from other scales are suppressed. We then place $F_{key}$ into the $F_{Fusion}^{2D}$ according to the position index of the $F_{key}$, obtaining the enhanced $F_{Fusion}^{2D^{'}} \in  R^{N_{voxels}\times C}$. The whole process of our adaptive fusion is illustrated in Fig.~\ref{fig:adaptivefusion}. 

\subsection{ Global Feature Aggregation}
% 改了开头
The LMFA module aims to learn the local contextual information around objects by dynamically aggregating neighborhood information of key voxels through the use of nearest neighbor positional relations.

Although the fusion of neighborhood voxel features enhances the expressive power of foreground sparse voxel features, LMFA module still has limitations when dealing with sparse detection scenarios. 1) For a large object, using a single aggregated sparse voxel as a proxy for object detection still suffers from information loss since the object proxy should contain information from the entire object rather than just a local region. 2) LMFA ignores the potential collaboration between the whole scene and instance features. For example, a false negative object in a scene can be potentially rectified by enhancing its feature through interactions with instances sharing similar semantic information. Therefore, we propose the GFA (Global Feature Aggregation) module to further address the limitations of the LMFA module by learning the global structural and semantic information of the entire scene. This makes SparseDet be able to leverage the contextual information of objects in both local and global ways to eliminate ambiguity, thereby improving the detection accuracy.

% LMFA 模块通过最近邻位置 relation 动态地聚合关键体素邻域information来增强其特征的表达能力。每个关键体素融合了一定范围内的邻域体素特征有效地增强了前景稀疏体素特征地表达能力，然而，LMFA模块仅可动态聚合局部范围的体素特征，在处理稀疏检测场景下仍存在限制：1）对于大物体来说，仅利用单个聚合好的稀疏体素作为物体检测的代理仍然面临着信息的损失，物体代理应包含整个物体的信息而不是局部。2）对于小物体以及远端物体来说，由于点云稀少以及稀疏化的表示，仅仅依靠局部信息可能无法准确检测到目标。因此，我们提出GFA模块来进一步解决LMFA模块的局限性。此外，我们的GFA模块避开了center feature特征缺失的问题。
% 全局上下文信息：点云数据表示了物体在三维空间中的几何结构，而全局 query可以捕捉整个点云的上下文信息。对于远端检测和小物体检测任务，目标点可能位于点云较远的区域或被其他物体遮挡，仅仅依靠局部信息可能无法准确检测到目标。通过全局 query，模型能够综合考虑整个点云的信息，包括点云的分布、密度和几何关系等，从而更好地定位和识别目标。
% 特征共享和全局交互：全局 query的方式可以在整个点云上共享特征和信息，从而提高模型的效率和性能。通过全局 query，模型可以将点云编码为一个统一的特征表示，使得模型可以更好地学习和推理。此外，全局 query还促进了点云中不同位置之间的交互和信息传递。这种交互可以帮助模型更好地理解点云中的语义和几何关系，提供更准确的目标检测结果。
% 尺度不变性和鲁棒性：全局 query可以提供尺度不变的特征表示。在点云中，目标的尺度可能有较大的变化，例如小物体可能只由少数几个点组成，或者远端物体可能只有少量稀疏的点。通过全局 query，模型可以对整个点云进行感知，从而获取尺度不变的特征表示。这使得模型能够更好地适应不同尺度的目标，提高检测的鲁棒性和泛化能力
% 在经过LMFA模块之后，我们得到了聚合好的关键体素特征F_key and 整个场景的体素特征 F_fusion, 其中 n_voels代表非空体素的数量。随后我们利用maxpool操作选择Score局部最大元素作为我们的对象查询。请注意，查询数量的设置为200。所选体素的位置和特征用于初始化查询位置和查询特征，as shown in 通过这种方式，我们最初的对象查询将位于或接近潜在对象中心，从而消除了对多个解码器层的需求来细化位置。将查询作为物体检测的代理天然的避免了中心特征缺失的问题。
%In the Local Multi-scale Feature Aggregation described above, we gained the aggregated features of the key voxels $F_{key}\in R^{N_{key}\times C}$ and the voxel features of the entire scene $F_{Fusion}^{2D^{'}} \in R^{N_{voxels}\times C}$, where $N_{voxels}$ respresents the number of non empty voxels. 

In detail, after LMFA module, we have the aggregated features of the key voxels $F_{key}\in R^{N_{key}\times C}$ and the enhanced voxel features of the entire scene $F_{Fusion}^{2D^{'}} \in R^{N_{voxels}\times C}$, where $N_{voxels}$ is the number of non empty voxels. Afterward, we select the top-$N_{query}$ highest-scoring voxels based on the $Score\in R^{N_{voxels}\times Cls}$ to initialize object queries. $N_{query}$ is set to 200 by default. We use the selected voxel positions $Pos_{query}$ from $\mathbb{R}^{N_{query} \times 2}$ and the features $F_{query}$ from $\mathbb{R}^{N_{query} \times C}$ to initialize the position encoding and the queries $Q$, as shown in Eq. (\ref{eq:query_init}). In this way, the initial object queries are located at or near potential object centers and contain more object information, enabling the queries to learn more efficiently and eliminating the need for multiple decoder layers to refine positions in query-based methods~\cite{transfusion,detr3d}.
\begin{equation}\label{eq:query_init}
    \begin{aligned}
        Q &= F_{query} + PE_{query} ,\\
        PE_{query} &= MLP(MLP(Pos_{query})).
    \end{aligned}
\end{equation}
% 在得到查询后，我们将稀疏体素特征F_fusion_2D特征作为K,V进行特征的聚合。然而由于场景中非空体素的数量并不固定，因此无法直接应用传统的decoder对稀疏物体查询进行特征聚合。我们统计了数据中，单个场景下的非空体素数量，发现范围大致在9K~13K之间
Where $PE_{query}$ means the position embedding of $Pos_{query}$. After obtaining the top-$N_{query}$ queries, we use the sparse voxel features $F_{Fusion}^{2D}$ as Key and Value in self-attention to obtain the attention score for each of top-$N_{query}$ queries so as to perform feature aggregation. However, due to the varying number of non-empty voxels in different scenes, traditional decoders cannot be directly applied to GFA sub-module. Thus, we have conducted data analysis and found that the number of non-empty voxels in a single scene ranges approximately from \textit{9k} to \textit{13k}. We then select a fixed number of $N_{K,V}$ sparse voxel features as the Key and the Value from the sparse voxel features $F_{Fusion}^{2D^{'}}$ based on $Score\in R^{N_{voxel}\times Cls}$. This allows SparseDet to focus more on foreground features. Here, $N_{K,V}$ is set to a default value of 10K, and any shortfall in the number of selected features is padded with zeros. 

In most approaches for voxel selection~\cite{voxelnext,transfusion}, it is common to flatten and sort $Score\in R^{N_{voxel}\times Cls}$ along the channel dimension. However, the aforementioned strategy cannot be directly applied in GFA module. The reason is that individual voxels often have similar scores for multiple classes, such as pedestrian and traffic cone. Relying solely on class scores as the selection criterion would result in the problem of repetitive feature selection. To address this issue, we propose a solution where we do not differentiate the scores for different classes. Instead, we consider the highest score within each voxel's class as the probability of that voxel being a foreground voxel. Specifically, we apply max-pooling along the channel dimension of $Score\in R^{N_{voxel}\times Cls}$ to obtain $Score{'}\in R^{N_{voxel}\times 1}$, which represents the highest score for each voxel regardless of its class. Next, we select $N_{K,V}$ sparse voxel features based on $Score{'}$ as the Key and the Value denoted as $F_{K}\in R^{N_{K,V}\times C}$ and $F_{V}\in R^{N_{K,V}\times C}$.

\begin{algorithm}[t]
\SetAlgoLined
\caption{The GFA Workflow} \label{algorithm:FBFusion}
\KwIn{

The height compressed sparse voxel features:   ${F_{Fusion}^{2D^{'}}}$\\
The heatmap of sparse voxels: $Score\in R^{N_{voxels}\times Cls}$\\
}
\KwOut{
The object queries: $\mathbf{Q}$\\
%Background Features   $\mathbf{F_{Back}}$

}
% \While{use image}{
%     \eIf{$\mathbf{F_{imp}}  > \mathcal{T}$}{ 
%         $\mathbf{F_{Fore}} =$
%     }

% }

\SetKwFunction{SASA}{SASA}%定义一个函数
\SetKwProg{Fn}{def}{\string :}{}% %定义python样式的函数格式

\Fn{\SASA{Q, $Pos_{query}$}}{%Python的类型注释写法
    $Dis = Calculate\_dis(Pos_{query})$\\
    $eta = MLP(Q)$\\
    $attn\_mask = Dis * eta$\\
    $Q = self\_attention(Q,attn\_mask)$\\
    Return $\mathbf{Q}$
}

\If{use GFA}{
    $top\_pro = Score.argsort()[..., : N_{query}]$\\
    $top\_pro\_cls = top\_pro // N_{voxels}$\\
    $top\_pro\_index = top\_pro \% Cls$\\
    $Q_{feat} =  F_{Fusion}^{2D^{'}}.features.gather($\\
    $\quad \quad \quad \quad \quad index=top\_pro\_index,$\\
    $\quad \quad \quad \quad \quad dim= -1) $\\
    $Pos_{query} = F_{Fusion}^{2D^{'}}.indices.gather($\\
    $\quad \quad \quad \quad \quad index=top\_pro\_index,$\\
    $\quad \quad \quad \quad \quad dim= -1) $\\
    $Score^{'} = Max\_pool(Score)$\\
    $top\_kv = Score^{'}.argsort()[..., : N_{K,V}]$\\
    $K,V_{feat} =  F_{Fusion}^{2D^{'}}.features.gather($\\
    $\quad \quad \quad \quad \quad index=top\_kv,$\\
    $\quad \quad \quad \quad \quad dim= -1) $\\
    $Pos_{K,V} = F_{Fusion}^{2D^{'}}.indices.gather($\\
    $\quad \quad \quad \quad \quad index=top\_kv,$\\
    $\quad \quad \quad \quad \quad dim= -1) $\\
    $Q = SASA(Q, Pos_{query})$\\
    $Q = decoder(Q_{feat}, K,V_{feat}, Pos_{query}, Pos_{K,V})$\\
     Return $\mathbf{Q}$
}

\end{algorithm}

% 尽管GFA模块具备全局的感受野，然而对于检测任务来说，并不需要时刻全局关注整个场景特征，我们希望在GFA保证全局感受野的同时，自适应地学习感受野更关注查询附近的体素特征。收到fastdetr的启发，我们为decoder引入尺度自适应自注意力，它在查询的指导下学习适当的感受野。首先, 我们计算查询之间的全部距离对。
In order to highlight the importance of target objects while maintaining a global perspective, GFA module should %while ensuring a global receptive field, 
interact more with the voxel features near the queries. Inspired from~\cite{fast_detr}, we introduce SASA (scale-adaptive self-attention) to GFA, allowing it to learn an appropriate receptive field guided by the queries.Firstly, %we compute the pairwise distances between all pairs of query centers. Firstly, 
SASA computes the pairwise distances $Dis\in R^{N_{query}\times N_{query}}$ between all pairs of query centers $Pos_{query}$ as below.

\begin{equation}\label{eq:dis}
    \begin{aligned}
        Dis_{i,j} = \sqrt{(X_{Pos}^{i}-X_{Pos}^{j})^{2} + (Y_{Pos}^{i}-Y_{Pos}^{j})^{2}},
    \end{aligned}
\end{equation}
% τ 用于控制每个查询的感受野，并由线性变换生成, 查询可以自适应地设置感受野的大小并更关注查询临近位置的特征。GFA模块的工作流程如算法所示。随后聚合好的Query特征被送入FFN层进行物体信息的解码
where ($X_{Pos}^{i}, Y_{Pos}^{i}$) denotes the positioin of the $i$-th query. Consequently, updated queries can be obtained in the following.
\begin{equation} \label{eq:adpative_fusion}
    \begin{aligned}
        Q = softmax(\frac{QK^{T}}{\sqrt{d} } + \eta log(Dis))V ,\\
    \end{aligned}
\end{equation}
where $\eta$ controls the receptive field of each query and is generated through a linear transformation. In this way, the queries $Q$ can dynamically adjust the size of receptive fields and prioritize features in the vicinity of the query centers. The workflow of the GFA module is detailed in Alg.~\ref{algorithm:FBFusion}. Afterwards, the aggregated and refined queries $Q$ are passed through a FFN (Feed Forward Network) layer for object prediction.

\section{Experiments}\label{Experiments}

In this section, we present the details of each dataset and the experimental setup of \textcolor{black}{SparseDet}, and evaluate SparseDet's performance of 3D object detection on KITTI~\cite{kitti} and nuScenes~\cite{nuscenes} datasets.

% %表11111111111111111111111111111111111
\subsection{Dataset and Evaluation Metrics}\label{sectionIV-A}
\subsubsection{KITTI dataset} 
The KITTI dataset~\cite{kitti} provides synchronized LiDAR point clouds and front-view camera images. It consists of 7,481 training samples and 7,518 test samples. \textcolor{black}{As a common practice \cite{voxelrcnn,focalconv,pvrcnn}, the training data are divided into a train set with 3712 samples and a validation set (\textit{val} set for short) with 3,769 samples to conduct evaluation. To perform evaluation on the test dataset using the official KITTI test server, we follow the approach outlined in VoxelNext~\cite{voxelnext}, use 80\% of the 7,481 training samples to train our model SparseDet with the amount of 5,985 samples.}
The standard evaluation metric for object detection is the mean Average Precision (mAP), computed using recall at 40 positions (R40). In this work, we evaluate SparseDet and all other dense and sparse methods on the most commonly used classes including \textcolor{black}{Car, Pedestrian and Cyclist} using Average Precision (AP) with an Intersection over Union (IoU) threshold of  \textcolor{black}{0.7, 0.5, and 0.5}, respectively. 

\subsubsection{nuScenes dataset} 
The nuScenes dataset~\cite{nuscenes} is a large-scale 3D detection benchmark consisting of 700 training scenes, 150 validation scenes, and 150 testing scenes. The dataset was collected using six multi-view cameras and a 32-beam LiDAR sensor. % and the dataset 
It includes 360-degree object annotations for 10 object classes. To evaluate the detection performance, the primary metrics used 
are the mean Average Precision (mAP) and \textcolor{black}{the nuScenes detection score (NDS)~\cite{nuscenes}, which assesses detection accuracy in terms of classification, bounding box location, size, orientation, attributes, and velocity.}
% For the ablation studies, we train models on subsets of the training data, including $\frac{1}{10}$, $\frac{1}{4}$, and the full dataset, and evaluate on the entire validation set, consisting of 75, 175, and 700 scenes for training and 150 scenes for validation. 

\begin{table*}[]
\scriptsize
\centering
\newcolumntype{M}[1]{>{\centering\arraybackslash}m{#1}}
\caption{Comparison with the SOTA methods on nuScenes test set. ``C.V.", ``Motor.", ``Ped.", and ``T.C." are short for construction vehicle, motorcycle, pedestrian, and traffic cone, respectively.}
\renewcommand\arraystretch{1}
\resizebox{\linewidth}{!}{
  %\begin{tabular*}{\linewidth} {@{}@{\extracolsep{\fill}}!{\color{white}\vline}l|c|c|c|c|c|c|c|c|c|c|c|c @{}}
\begin{tabular}{l|l|c|c|c|c|c|c|c|c|c|c|c|c }
\toprule
\multicolumn{2}{c|}{Method}   & mAP  & NDS  & Car  & Truck & C.V. & Bus  & Trailer & Barrier & Motor. & Bike & Ped. & T.C. \\ 
\midrule
\multirow{12}{*}{Dense} & PointPillars\cite{pointpillars} & 30.5& 45.3 & 68.4 &23.0& 4.1 &28.2 &23.4 &38.9 &27.4& 1.1& 59.7& 30.8\\
& 3DSSD\cite{3dssd} & 42.7 & 56.4 & 81.2 & 47.2 & 12.6 & 61.4 & 30.5 & 47.9 & 36.0 & 8.6 & 70.2 & 31.1\\
& SASA\cite{SASA} & 45.0 & 61.0 & 76.8 & 45.0 & 16.1 & 66.2 & 36.5 & 53.6 & 39.6 & 16.9 & 69.1 & 29.9 \\
& CenterPoint\cite{centerpoint} & 58.0 & 65.5 & 84.6 & 51.0 & 17.5 & 60.2 & 53.2 & 70.9 & 53.7 & 28.7 & 83.4 & 76.7\\
& HotSpotNet\cite{Hotspots} & 59.3 & 66.0 & 83.1 & 50.9 & 23.0 & 56.4 & 53.3 & 71.6 & 63.5 & 36.6 & 81.3 & 73.0\\
& InfoFocus  \cite{infofocus}               & 39.5 & 39.5 & 77.9 & 31.4  & 10.7 & 44.8 & 37.3    & 47.8    & 29.0   & 6.1  & 63.4 & 46.5 \\
& AFDetV2  \cite{afdetv2}            & 62.4 & 68.5 & 86.3 & 54.2  & 26.7 & 62.5 & 58.9    & 71.0    & 63.8   & 34.3  & 85.8 & 80.1 \\
& VISTA  \cite{vista}            & 63.0 & 69.8 & 84.4 & 55.1  & 25.1 & 63.7 & 54.2    & 71.4    & 70.0   & 45.4  & 82.8 & 78.5 \\
& Focals Conv\cite{focalconv} & 63.8& 70.0 & 86.7 & 56.3 & 23.8 & 67.7 & 59.5 & 74.1 & 64.5 & 36.3 & 81.4& 81.4\\
& TransFusion-L \cite{transfusion} & 65.5 & 70.2 & 86.2 & 56.7 & 28.2 & 66.3 & 58.8 & 78.2 & 68.3 & 44.2 & 86.1 & 82.0\\
& UVTR\cite{uvtr} &67.1& 71.1& 87.5& 56.0& 33.8& 67.5& 59.5 &73.0& 73.4 &54.8 &86.3& 79.6 \\
& LargeKernel3D\cite{largekernel3d} & 65.3 & 70.5 & 85.9 & 55.3 & 26.8 & 65.7 & 62.1 & 75.5 & 72.5 & 46.6 & 85.6 & 80.0 \\
& PVT-SSD\cite{pvt}  & 53.6 & 65.0 & 79.4 & 43.8 & 21.7 & 62.1 & 34.2 & 67.1 & 53.4 & 38.2 & 79.8 & 56.6 \\
\midrule
\multirow{4}{*}{Sparse} & FSD \cite{FSD}                  & 62.5 & 68.7 & - & - & - & - & - & - & - & - & - & -\\
& FSDV2 \cite{FSDV2}                  & 66.2 & 71.7 & 83.7 & 51.6 & \textbf{32.5} & 66.4 & \textbf{59.1} & \textbf{78.7} & 71.4 & \textbf{51.7} & 87.1 & 80.3\\
\cmidrule(lr){2-14}
& VoxelNeXt\cite{voxelnext} & 64.5& 70.0 &84.6 &53.0& 28.7 &64.7 &55.8 &74.6 &73.2& 45.7& 85.8& 79.0 \\
& \cellcolor{blue!10} \textbf{\textcolor{black}{SparseDet}}  & \cellcolor{blue!10} \textbf{66.7}\textit{\fontsize{6}{0}\selectfont\textcolor{red}{+2.2}} & \cellcolor{blue!10} \textbf{71.9}\textit{\fontsize{6}{0}\selectfont\textcolor{red}{+1.9}} & \cellcolor{blue!10} \textbf{86.2}\textit{\fontsize{6}{0}\selectfont\textcolor{red}{+1.6}} & \cellcolor{blue!10} \textbf{56.0}\textit{\fontsize{6}{0}\selectfont\textcolor{red}{+3.0}} & \cellcolor{blue!10} 30.2\textit{\fontsize{6}{0}\selectfont\textcolor{red}{+1.5}} & \cellcolor{blue!10} \textbf{66.5}\textit{\fontsize{6}{0}\selectfont\textcolor{red}{+1.8}} & \cellcolor{blue!10} 58.4\textit{\fontsize{6}{0}\selectfont\textcolor{red}{+2.6}} & \cellcolor{blue!10} 78.7\textit{\fontsize{6}{0}\selectfont\textcolor{red}{+4.1}} & \cellcolor{blue!10} \textbf{73.7}\textit{\fontsize{6}{0}\selectfont\textcolor{red}{+0.5}} & \cellcolor{blue!10} 46.8\textit{\fontsize{6}{0}\selectfont\textcolor{red}{+1.1}} & \cellcolor{blue!10} \textbf{87.5}\textit{\fontsize{6}{0}\selectfont\textcolor{red}{+1.7}} & \cellcolor{blue!10} \textbf{82.5}\textit{\fontsize{6}{0}\selectfont\textcolor{red}{+3.5}}  \\
\bottomrule
\end{tabular}
}
\label{tab_nuScens_test}
\end{table*}

\begin{table*}[tbh]
\scriptsize
\centering
\caption{Comparison with the baseline on nuScenes \textit{val} dataset. ‘C.V.’, ‘Ped.’, and ‘T.C.’ are short for construction vehicle, pedestrian, and traffic cone, respectively. }
\renewcommand\arraystretch{1}
\resizebox{\linewidth}{!}{
\begin{tabular}{c|c|c|c|c|c|c|c|c|c|c|c|c|c}
\toprule
Dataset Split & Method & mAP & NDS & Car & Truck & C.V. & Bus & Trailer & Barrier & Motor. & Bike & Ped. & T.C. \\ 
\hline
\multirow{2}{*}{full} & VoxelNeXt$^*$ & 60.8 & 68.1 & 83.8 & 57.7 & 20.8 & 71.5 & 38.6 & 67.6 & 63.0 & 51.5 & 84.5 & 69.0 \\
& \cellcolor{blue!10} \textbf{\textcolor{black}{SparseDet}} & \cellcolor{blue!10}  65.3\textit{\fontsize{6}{0}\selectfont\textcolor{red}{+4.5}} & \cellcolor{blue!10} 70.3\textit{\fontsize{6}{0}\selectfont\textcolor{red}{+2.2}} & \cellcolor{blue!10} 87.5\textit{\fontsize{6}{0}\selectfont\textcolor{red}{+3.7}} & \cellcolor{blue!10} 60.2\textit{\fontsize{6}{0}\selectfont\textcolor{red}{+2.5}} & \cellcolor{blue!10} 27.2\textit{\fontsize{6}{0}\selectfont\textcolor{red}{+6.4}} & \cellcolor{blue!10} 75.8\textit{\fontsize{6}{0}\selectfont\textcolor{red}{+4.3}} & \cellcolor{blue!10} 40.4\textit{\fontsize{6}{0}\selectfont\textcolor{red}{+1.8}} & \cellcolor{blue!10} 73.1\textit{\fontsize{6}{0}\selectfont\textcolor{red}{+5.5}} & \cellcolor{blue!10} 69.7\textit{\fontsize{6}{0}\selectfont\textcolor{red}{+6.7}} & \cellcolor{blue!10} 58.8\textit{\fontsize{6}{0}\selectfont\textcolor{red}{+7.3}} & \cellcolor{blue!10} 86.5\textit{\fontsize{6}{0}\selectfont\textcolor{red}{+2.0}} & \cellcolor{blue!10} 63.4\textit{\fontsize{6}{0}\selectfont\textcolor{red}{+4.4}} \\  
\hline
\multirow{2}{*}{$\frac{1}{4} $} & VoxelNeXt$^*$ & 51.8 & 60.7 & 80.2 & 48.1 & 15.4 & 63.1 & 26.1 & 59.3 & 52.5 & 35.5 & 81.6 & 56.2 \\
& \cellcolor{blue!10} \textbf{\textcolor{black}{SparseDet}} & \cellcolor{blue!10}  55.3\textit{\fontsize{6}{0}\selectfont\textcolor{red}{+3.5}} & \cellcolor{blue!10} 62.7\textit{\fontsize{6}{0}\selectfont\textcolor{red}{+2.0}} & \cellcolor{blue!10} 82.7\textit{\fontsize{6}{0}\selectfont\textcolor{red}{+2.5}} & \cellcolor{blue!10} 51.8\textit{\fontsize{6}{0}\selectfont\textcolor{red}{+3.7}} & \cellcolor{blue!10} 18.8\textit{\fontsize{6}{0}\selectfont\textcolor{red}{+3.4}} & \cellcolor{blue!10} 64.3\textit{\fontsize{6}{0}\selectfont\textcolor{red}{+1.2}} & \cellcolor{blue!10} 27.6\textit{\fontsize{6}{0}\selectfont\textcolor{red}{+1.5}} & \cellcolor{blue!10} 64.6 \textit{\fontsize{6}{0}\selectfont\textcolor{red}{+5.3}}& \cellcolor{blue!10} 60.1\textit{\fontsize{6}{0}\selectfont\textcolor{red}{+7.6}} & \cellcolor{blue!10} 39.1\textit{\fontsize{6}{0}\selectfont\textcolor{red}{+3.6}} & \cellcolor{blue!10} 81.6 & \cellcolor{blue!10} 62.7\textit{\fontsize{6}{0}\selectfont\textcolor{red}{+6.5}} \\ 
\hline
\multirow{2}{*}{$\frac{1}{8}$} & VoxelNeXt$^*$ & 46.6 & 55.4 & 78.0 & 41.8 & 15.9 & 57.7 & 20.6 & 53.1 & 49.4 & 23.7 & 76.3 & 49.4 \\
& \cellcolor{blue!10} \textbf{\textcolor{black}{SparseDet}} & \cellcolor{blue!10}  49.5\textit{\fontsize{6}{0}\selectfont\textcolor{red}{+2.9}} & \cellcolor{blue!10} 58.2\textit{\fontsize{6}{0}\selectfont\textcolor{red}{+2.8}} & \cellcolor{blue!10} 81.3\textit{\fontsize{6}{0}\selectfont\textcolor{red}{+3.3}} & \cellcolor{blue!10} 44.8\textit{\fontsize{6}{0}\selectfont\textcolor{red}{+3.0}} & \cellcolor{blue!10} 17.4\textit{\fontsize{6}{0}\selectfont\textcolor{red}{+1.5}} & \cellcolor{blue!10} 63.2\textit{\fontsize{6}{0}\selectfont\textcolor{red}{+5.5}} & \cellcolor{blue!10} 24.1\textit{\fontsize{6}{0}\selectfont\textcolor{red}{+3.5}} & \cellcolor{blue!10} 54.2\textit{\fontsize{6}{0}\selectfont\textcolor{red}{+1.1}} & \cellcolor{blue!10} 54.2\textit{\fontsize{6}{0}\selectfont\textcolor{red}{+4.8}} & \cellcolor{blue!10} 25.7\textit{\fontsize{6}{0}\selectfont\textcolor{red}{+2.0}} & \cellcolor{blue!10} 78.8\textit{\fontsize{6}{0}\selectfont\textcolor{red}{+2.5}} & \cellcolor{blue!10} 52.6\textit{\fontsize{6}{0}\selectfont\textcolor{red}{+3.2}} \\
\hline
\end{tabular}
}
\label{tab_nuScenes_val}
\begin{tablenotes}
\footnotesize
\item[1] * denotes re-implement result.
\item[2] The color \textcolor{red}{red} indicates improvement.
\end{tablenotes}
\end{table*}

\begin{table*}[t]
% \small% 设置字体大小命令由小到大依次为 \tiny \scriptsize \footnotesize \small
% \footnotesize
\scriptsize
\centering
\addtolength{\tabcolsep}{1.3pt}
\caption{Performance comparison with SOTA methods on {KITTI test} set for 3D detection with an average precision of 40 sampling recall points evaluated on KITTI server. ‘L’ represents LiDAR.}
\renewcommand\arraystretch{1}
\setlength{\tabcolsep}{0.95mm}{
\begin{tabular*}{\linewidth}{l|l|c|c|c|c|c|c|c|c|c|c|c|c|c|c|c|c|c|c}
\toprule
\multicolumn{2}{c|}{}       & \multicolumn{6}{c|}{Car}                                                                                                                       & \multicolumn{6}{c|}{Pedestrian}                                                                                                                & \multicolumn{6}{c}{Cyclist}                                                                                                                    \\ \cmidrule(lr){3-20} 
\multicolumn{2}{c|}{Method}  & \multicolumn{3}{c|}{$AP_{3D}$ (\%)}                                                        & \multicolumn{3}{c|}{$AP_{BEV}$ (\%)}                                  & \multicolumn{3}{c|}{$AP_{3D}$ (\%)}                                                        & \multicolumn{3}{c|}{$AP_{BEV}$ (\%)}                                  & \multicolumn{3}{c|}{$AP_{3D}$ (\%)}                                                        & \multicolumn{3}{c}{$AP_{BEV}$ (\%)}                                   \\ 
\cmidrule(lr){3-20} 
\multicolumn{2}{c|}{}      & \multicolumn{1}{l|}{Easy} & \multicolumn{1}{l|}{Mod} & \multicolumn{1}{l|}{Hard} & \multicolumn{1}{c|}{Easy} & \multicolumn{1}{c|}{Mod} & Hard & \multicolumn{1}{c|}{Easy} & \multicolumn{1}{c|}{Mod} & \multicolumn{1}{c|}{Hard} & \multicolumn{1}{c|}{Easy} & \multicolumn{1}{c|}{Mod} & Hard & \multicolumn{1}{c|}{Easy} & \multicolumn{1}{c|}{Mod} & \multicolumn{1}{c|}{Hard} & \multicolumn{1}{c|}{Easy} & \multicolumn{1}{c|}{Mod} & Hard \\ \midrule

\multirow{13}{*}{Dense} & BSAODet \cite{xiao2023balanced}                                         & \multicolumn{1}{c|}{88.89}          & \multicolumn{1}{c|}{81.74}          & 77.14           & \multicolumn{1}{c|}{-}          & \multicolumn{1}{c|}{-}          & -      & 51.71    & 43.63   & 41.09   & -    & -   & -    & \textbf{82.65}    & \textbf{67.79}   & 60.26    & -    & -   & -    \\
&  H$^2$3D R-CNN \cite{deng2021multi}                                       & \multicolumn{1}{c|}{90.43}          & \multicolumn{1}{c|}{81.55}          & 77.22           & \multicolumn{1}{c|}{92.85}          & \multicolumn{1}{c|}{88.87}          & 86.07    & 52.75    & 45.26   & 41.56    & 58.14    & 50.43   & 46.72    & 78.67    & 62.74   & 55.78    & 82.76    & 67.90   & 60.49      \\
&  SECOND \cite{SECOND}                                        & \multicolumn{1}{c|}{84.65}          & \multicolumn{1}{c|}{75.96}          & 68.71          & \multicolumn{1}{c|}{91.81}          & \multicolumn{1}{c|}{86.37}          & 81.04 & -    & -   & -    & -    & -   & -    & -    & -   & -    & -    & -   & -         \\
&  PointPillars \cite{pointpillars}                               & \multicolumn{1}{c|}{82.58}          & \multicolumn{1}{c|}{74.31}          & 68.99          & \multicolumn{1}{c|}{90.07}          & \multicolumn{1}{c|}{86.56}          & 82.81   & \multicolumn{1}{c|}{51.45}    & \multicolumn{1}{c|}{41.92}   & \multicolumn{1}{c|}{38.89}    & \multicolumn{1}{c|}{57.60}    & \multicolumn{1}{c|}{48.64}   & 45.78    & \multicolumn{1}{c|}{77.10}    & \multicolumn{1}{c|}{58.65}   & \multicolumn{1}{c|}{51.92}    & \multicolumn{1}{c|}{79.90}    & \multicolumn{1}{c|}{62.73}   & 55.58       \\
&SIEV-Net \cite{SIENet} & \multicolumn{1}{c|}{85.21} & \multicolumn{1}{c|}{76.18} & \multicolumn{1}{c|}{70.60} & - & - & - & \multicolumn{1}{c|}{54.00} & \multicolumn{1}{c|}{44.80} & \multicolumn{1}{c|}{41.11} & -  & -  & -  & \multicolumn{1}{c|}{78.75} & \multicolumn{1}{c|}{59.99} & \multicolumn{1}{c|}{52.37} & - & - & -\\

&  VoxSet \cite{voxset}                                   & \multicolumn{1}{c|}{88.53}          & \multicolumn{1}{c|}{82.06}          & 77.46          & \multicolumn{1}{c|}{-}              & \multicolumn{1}{c|}{-}              & -      & \multicolumn{1}{c|}{-}    & \multicolumn{1}{c|}{-}   & \multicolumn{1}{c|}{-}    & \multicolumn{1}{c|}{-}    & \multicolumn{1}{c|}{-}   & -    & \multicolumn{1}{c|}{-}    & \multicolumn{1}{c|}{-}   & \multicolumn{1}{c|}{-}    & \multicolumn{1}{c|}{-}    & \multicolumn{1}{c|}{-}   & -        \\
&  TANet \cite{tanet}                                    & \multicolumn{1}{c|}{83.81}          & \multicolumn{1}{c|}{75.38}          & 67.66          & \multicolumn{1}{c|}{-}          & \multicolumn{1}{c|}{-}          & -    & \multicolumn{1}{c|}{\textbf{54.92}}    & \multicolumn{1}{c|}{46.67}   & \multicolumn{1}{c|}{\textbf{42.42} }    & \multicolumn{1}{c|}{-}    & \multicolumn{1}{c|}{-}   & -    & \multicolumn{1}{c|}{73.84}    & \multicolumn{1}{c|}{59.86}   & \multicolumn{1}{c|}{53.46}    & \multicolumn{1}{c|}{-}    & \multicolumn{1}{c|}{-}   & -       \\
&  Part-A$^{2}$ \cite{parta2}                                 & \multicolumn{1}{c|}{87.81}          & \multicolumn{1}{c|}{78.49}          &  73.51          & \multicolumn{1}{c|}{91.70}          & \multicolumn{1}{c|}{ 87.79}          & 84.61    & \multicolumn{1}{c|}{53.10}    & \multicolumn{1}{c|}{43.35}   & \multicolumn{1}{c|}{40.06}    & \multicolumn{1}{c|}{59.04}    & \multicolumn{1}{c|}{49.81}   & 45.92    & \multicolumn{1}{c|}{79.17}    & \multicolumn{1}{c|}{63.52}   & \multicolumn{1}{c|}{56.93}    & \multicolumn{1}{c|}{\textbf{83.43}}    & \multicolumn{1}{c|}{68.73}   & 61.85      \\
&  Voxel RCNN \cite{voxelrcnn}                            & \multicolumn{1}{c|}{\textbf{90.90}}          & \multicolumn{1}{c|}{81.62}          & 77.06          & \multicolumn{1}{c|}{-}              & \multicolumn{1}{c|}{-}              & -       & \multicolumn{1}{c|}{-}    & \multicolumn{1}{c|}{-}   & \multicolumn{1}{c|}{-}    & \multicolumn{1}{c|}{-}    & \multicolumn{1}{c|}{-}   & -    & \multicolumn{1}{c|}{-}    & \multicolumn{1}{c|}{-}   & \multicolumn{1}{c|}{-}    & \multicolumn{1}{c|}{-}    & \multicolumn{1}{c|}{-}   & -         \\
&  Voxel RCNN *                              & \multicolumn{1}{c|}{90.76}          & \multicolumn{1}{c|}{81.69}          & 77.42          & \multicolumn{1}{c|}{92.89}          & \multicolumn{1}{c|}{89.97}          & 84.69      & \multicolumn{1}{c|}{52.57}    & \multicolumn{1}{c|}{44.86}   & \multicolumn{1}{c|}{39.09}    & \multicolumn{1}{c|}{57.66}    & \multicolumn{1}{c|}{49.32}   & 44.15    & \multicolumn{1}{c|}{77.54}    & \multicolumn{1}{c|}{64.00}   & \multicolumn{1}{c|}{53.15}    & \multicolumn{1}{c|}{79.68}    & \multicolumn{1}{c|}{67.56}   & 62.70      \\
&  PV-RCNN \cite{pvrcnn}                                 & \multicolumn{1}{c|}{90.25}          & \multicolumn{1}{c|}{81.43}          & 76.82          & \multicolumn{1}{c|}{94.98}          & \multicolumn{1}{c|}{90.65}          & 86.14    & 52.17     & 43.29      & 40.29   & 59.86     & 50.57     & 46.74     & 78.60     & 63.71     & 57.65     & 82.49         & 68.89     & 62.41    \\
&  PV-RCNN *                                  & \multicolumn{1}{c|}{90.61}          & \multicolumn{1}{c|}{81.51}          & 76.81          & \multicolumn{1}{c|}{94.68}          & \multicolumn{1}{c|}{90.87}          & 86.19    & 52.10    & 43.63      & 40.44   & 60.06     & 50.43     & 46.81     & 78.58     & 63.83     & 57.71     & 82.50         & 68.93     & 62.57    \\
&  PV-RCNN++ *\cite{PV-RCNN++} & 87.72 & 81.29 & 76.78 & 91.93 & 89.75 & 85.77 & 47.50 & 40.31 & 38.15 & 51.20 & 44.61 & 42.55 & 80.34 & 67.46 & 60.38 & 83.83 & 71.51 & 64.77\\
&  PVT-SSD\cite{pvt} & 90.65 & \textbf{82.29} & 76.85 & \textbf{95.23} & \textbf{91.63} & 86.43 & - & - & - & - & - & - & - & - & - & - & - & -\\
\midrule
\multirow{2}{*}{Sparse} & VoxelNeXt *\cite{voxelnext} & 89.10 & 80.35 & 76.99 & 93.03 & 89.49 & 86.40 & 52.10 & 42.72 & 39.08 & 57.19 & 48.27 & 44.55 & 81.33 & 65.31 & 57.43 & 83.34 & 67.55 & 59.61  \\
& \cellcolor{blue!10} \textbf{SparseDet} & \cellcolor{blue!10} 90.79 & \cellcolor{blue!10} 81.17 & \cellcolor{blue!10} \textbf{78.11} & \cellcolor{blue!10} 94.48 & \cellcolor{blue!10} 89.68 & \cellcolor{blue!10} \textbf{87.53} & \cellcolor{blue!10} 52.92 & \cellcolor{blue!10} 44.61 & \cellcolor{blue!10} 41.80 & \cellcolor{blue!10} 59.84 & \cellcolor{blue!10} 50.75 & \cellcolor{blue!10} 44.57 & \cellcolor{blue!10} 81.93 & \cellcolor{blue!10} 65.95 & \cellcolor{blue!10} \textbf{60.41} & \cellcolor{blue!10} 83.16 & \cellcolor{blue!10} \textbf{69.04} & \cellcolor{blue!10} \textbf{63.46}\\
\bottomrule
\end{tabular*}
\label{tab_kitti_test}
\begin{tablenotes}
\footnotesize
\item[1] * denotes re-implement result.
\end{tablenotes}
}
\end{table*}

\begin{table}[]
\scriptsize
\centering
\caption{Performance comparison with the SOTA methods on {KITTI \textit{val}} dataset for car class. The results are reported by the mAP with 0.7 IoU threshold and 40 recall points. ‘L’ represents LiDAR.}
\renewcommand\arraystretch{1}
\setlength{\tabcolsep}{2.4mm}{

\begin{tabular}{@{}@{\extracolsep{\fill}}!{\color{white}\vline}l|c|c|c|c|c|c @{}}
\toprule
\multirow{2}{*}{Method} & \multicolumn{3}{c|}{$AP_{3D}$ (\%)}   & \multicolumn{3}{c}{$AP_{BEV}$ (\%)}                                                                                                                      \\ \cmidrule(r){2-7} 
                           & Easy           & Mod.           & Hard    & Easy           & Mod.           & Hard            \\ \midrule
PointRCNN \cite{pointrcnn}                         & 88.88          & 78.63          & 77.38    & -             & -             & -           \\
H$^2$3D R-CNN \cite{deng2021multi}                        & 89.63          & 85.20          & 79.08    & -             & -             & -           \\

SECOND \cite{SECOND}                         & 87.43         & 76.48          & 69.10   & -             & -             & -               \\
MedTr-TSD \cite{tian2023medoidsformer}                         &  89.27        &  84.24          & 78.85   & -            & -            & -                \\
CT3D \cite{ct3d}                         & 92.85 & \textbf{85.82}          & 83.46    & \textbf{96.14}	& 91.88	 & 89.63
          \\

Part-A$^{2}$ \cite{parta2}                         & 89.47          & 79.47          & 78.54    & 90.42          & 88.61          & 87.31               \\
Voxel RCNN \cite{voxelrcnn}                         & 92.38          & 85.29          & 82.86   & 95.52          & 91.25          & 88.99            \\
PV-RCNN \cite{pvrcnn}                         & 92.57          & 84.83          & 82.69   & 95.76 & 91.11          & 88.93              \\
PC-RGNN\cite{PC-RGNN} & 90.94 & 81.43 & 80.45 & - & - & - \\
PDV\cite{PDV} & 92.56 & 85.29 & 83.05 & - & - & - \\
SIENet\cite{SIENet} & 91.96 & 84.45 & 82.64 & - & - & -\\
\midrule        
%VoxelNeXt *\cite{voxelnext}                         & 90.43          & 81.43          & 78.44   & 94.15          & 90.13          & 87.70            \\
VoxelNeXt*\cite{voxelnext}  & 92.51     & 84.35 & 82.71  & 95.24  & 90.28   & 90.45 \\ 
\rowcolor{blue!10}SparseDet                      & \textbf{93.81}     & 84.78 & \textbf{84.33}  & 95.30  & \textbf{92.55}   & \textbf{91.02}  \\ 
\bottomrule
\end{tabular}
\begin{tablenotes}
\footnotesize
\item[1] * denotes re-implement result.
\end{tablenotes}
}
\label{tab_kitti_val}
\end{table}

% flops Mem FPS KNN (kitti focals conv) 不同距离

\begin{table}[t]
\scriptsize
\centering
\caption{Runtime comparison on nuScenes\cite{nuscenes} dataset. Mem. denotes the training GPU memory measured following\cite{SECOND}. FPS (frame per second) is the inference speed measured on a single NVIDIA 3090 GPU with a batch size of 1.}
\renewcommand\arraystretch{1}
%\begin{tabular*}{\linewidth}{@{\extracolsep{\fill}}!{\color{white}\vline}c|c|c|c|c|c|c}
%\tabcolsep=3.3mm
\resizebox{\linewidth}{!}{
\begin{tabular}{l|c|c|c|c|c}
\toprule
Method & mAP & NDS  & FPS  & \#Params & Mem.(G) \\ 
\midrule
FSD\cite{FSD} & 62.5 & 68.7 & 10.1 & 11.7M & 4.97\\
FSDV2\cite{FSDV2}& 66.2 & 71.7 & 10.3 & 10.6M & 4.80\\
VoxelNeXt\cite{voxelnext}  &  64.5  &   70.0  &   15.1   & 7.62M & 3.42    \\
\rowcolor{blue!10} SparseDet & 66.7 & 71.9 & 13.5 & 7.98M & 3.71 \\
\bottomrule                            
\end{tabular}}
\label{tab_sparse_compare}
\end{table}

\subsection{Implementation Details}

In this section, we will provide a detailed description of the SparseDet settings for KITTI~\cite{kitti} and nuScenes~\cite{nuscenes}. To enable effective training on KITTI and nuScenes, we utilize 8 NVIDIA RTX 3090 24G GPUs. During inference, the batch size is set to 1 on a 3090 GPU. All latency measurements are taken on the same workstation
with a 3090 GPU. We implement SparseDet based on the open-source code repositories OpenPCDet~\cite{openpcdet} and the baseline VoxelNeXt~\cite{voxelnext}. 

For KITTI~\cite{kitti}, the input voxel size is set to [0.05m, 0.05m, 0.1m] and the point range is set to [0m, -40m, -3m, 70.4m, 40m, 1m] across the X, Y and Z axes, respectively. The maximum number of point clouds contained in each voxel is set to 10. During training, following the baseline~\cite{voxelnext}, SparseDet is trained 80 epoches.

For nuScenes~\cite{nuscenes}, the input voxel size is set to [0.075m, 0.075m, 0.2m] and the point range is set to [-54m, -54m, -5m, 54m, 54m, 3m] across the X, Y and Z axes, respectively. During training, following the baseline~\cite{voxelnext}, SparseDet is trained 20 epoches. For more details concerning our method, please refer to OpenPCDet~\cite{openpcdet}.

\begin{table}[t]
\scriptsize
\centering
\caption{Effect of each component in our \textcolor{black}{SparseDet}. Results are reported on KITTI \textit{val} set for \textcolor{black}{car category} with VoxelNeXt. FPS (frame per second) is the inference speed measured on a single NVIDIA 3090 GPU with a batch size of 1.}

  \renewcommand\arraystretch{1}
  %\begin{tabular*}{\linewidth}{@{\extracolsep{\fill}}!{\color{white}\vline}c|c|c|c|c|c|c}
  %\tabcolsep=2.95mm
  \resizebox{\linewidth}{!}{
  \begin{tabular}{c|c|c|c|c|c}
\toprule
\multirow{3}{*}{LMFA} & \multirow{3}{*}{GFA} & \multicolumn{2}{c|}{Hard}                                        & \multirow{3}{*}{\#Params} & \multirow{3}{*}{FPS} \\
\cmidrule(lr){3-4}
                      &                                  & $AP_{3D}$(\%)           & $AP_{BEV}$(\%)                     &                                   &                       \\ 
                        \midrule
                      &                       & 78.44 & 87.10       & 7.45 M                           & 20.1                  \\
 \checkmark                     &                                  & 81.67\textit{\fontsize{6}{0}\selectfont\textcolor{red}{+3.23}}  &  89.24\textit{\fontsize{6}{0}\selectfont\textcolor{red}{+2.14}}  & 7.55 M                           & 18.6                  \\
\rowcolor{blue!10}\checkmark                     & \checkmark                                & 84.33\textit{\fontsize{6}{0}\selectfont\textcolor{red}{+2.66}}  &  91.02\textit{\fontsize{6}{0}\selectfont\textcolor{red}{+1.78}}         & 7.79 M                           & 17.9                  \\ 
     \bottomrule
\end{tabular}}
\label{tab_ablation_kitti_voxel}
\begin{tablenotes}
\footnotesize
\item[1] The color \textcolor{red}{red} indicates improvement.
\end{tablenotes}
\end{table}

\begin{table}[t]
\scriptsize
\centering
\caption{Effect of each component in our \textcolor{black}{SparseDet}. Results are reported on nuScenes \textit{val} set (trained on $\frac{1}{4}$ subset) with VoxelNeXt. FPS (frame per second) is the inference speed measured on a single NVIDIA 3090 GPU with a batch size of 1.}
\renewcommand\arraystretch{1}
\resizebox{\linewidth}{!}{
\begin{tabular}{c|c|c|c|c|c}
\toprule
LMFA & GFA & mAP  & NDS  & \#Params & FPS \\ \midrule
   &   & 51.8  & 60.7  &  7.62M   &   15.1  \\
 \checkmark     &    & 52.9\textit{\fontsize{6}{0}\selectfont\textcolor{red}{+1.1}} & 61.4\textit{\fontsize{6}{0}\selectfont\textcolor{red}{+0.7}} & 7.71M  &   14.0 \\
\rowcolor{blue!10}   \checkmark    &  \checkmark  & 55.3\textit{\fontsize{6}{0}\selectfont\textcolor{red}{+2.4}}  &  62.7\textit{\fontsize{6}{0}\selectfont\textcolor{red}{+1.3}} & 7.98M  &  13.5 \\ 
\bottomrule                             
\end{tabular}}
\label{tab_ablation_nuscens}
\begin{tablenotes}
\footnotesize
\item[1] The color \textcolor{red}{red} indicates improvement. 
\end{tablenotes}
\end{table}

\begin{table}[t]
\scriptsize
\centering
\caption{Effect of the number of $M$ on  nuScenes \textit{val} set (trained on $\frac{1}{4}$ subset)  with SparseDet.}
\renewcommand\arraystretch{1}
%\begin{tabular*}{\linewidth}{@{\extracolsep{\fill}}!{\color{white}\vline}c|c|c|c|c|c|c}
%\tabcolsep=3.5mm
\resizebox{\linewidth}{!}{
\begin{tabular}{c|c|c|c|c|c}
\toprule
$M$ & mAP& NDS & FPS & \#params & Mem.(G)\\ \midrule
4   & 54.7   & 62.5 & 13.8 & 7.92M & 3.60\\
8  & \textbf{55.3}   & 62.7 & 13.5 & 7.98M & 3.71\\
16  & 55.2   & \textbf{62.9} & 12.9 & 8.11M& 3.88\\
32  & 54.3   & 62.1 & 12.1 & 8.37M & 4.01\\
 \midrule                             
\end{tabular}}
\label{tab_ablation_nus_M}
\end{table}

\begin{table}[t]
\scriptsize
\centering
\caption{Effect of the number of $N_{key}$ on  nuScenes \textit{val} set (trained on $\frac{1}{4}$ subset)  with SparseDet.}
\renewcommand\arraystretch{1}
%\begin{tabular*}{\linewidth}{@{\extracolsep{\fill}}!{\color{white}\vline}c|c|c|c|c|c|c}
%\tabcolsep=3.5mm
\resizebox{\linewidth}{!}{
\begin{tabular}{c|c|c|c|c|c}
\toprule
$N_{key}$ & mAP& NDS & FPS & \#params & Mem.(G)\\ \midrule
500   & 55.3   & 62.7 & 13.5 & 7.98M & 3.71\\
1000  & 55.6   & \textbf{63.2}  & 12.9 & 7.98M & 3.89\\
1500  & 55.6   & 62.6 & 12.0 & 7.98M & 3.93\\
2000  & \textbf{55.8}   & 63.0 & 11.6 & 7.98M & 3.98\\
 \midrule                             
\end{tabular}}
\label{tab_ablation_nus_N_Key}
\end{table}

\begin{table}[t]
\scriptsize
\centering
\caption{Effect of the number of $N_{K,V}$ on  nuScenes \textit{val} set (trained on $\frac{1}{4}$ subset)  with SparseDet.}
\renewcommand\arraystretch{1}
%\begin{tabular*}{\linewidth}{@{\extracolsep{\fill}}!{\color{white}\vline}c|c|c|c|c|c|c}
%\tabcolsep=3.5mm
\resizebox{\linewidth}{!}{
\begin{tabular}{c|c|c|c|c|c}
\toprule
$N_{K,V}$ & mAP& NDS & FPS & \#params & Mem.(G)\\ \midrule
6000   & 54.1   & 61.7 & 14.2& 7.98M & 3.51\\
8000  & 54.9   & 62.3 & 13.8 & 7.98M & 3.62\\
10000  & 55.3   & 62.7 & 13.5 & 7.98M & 3.71\\
12000  & \textbf{55.4}   & \textbf{62.9} & 13.0 & 7.98M & 3.83\\
 \midrule                             
\end{tabular}}
\label{tab_ablation_nus_NKV}
\end{table}

\subsection{Comparison with State-of-the-Arts}
\subsubsection{Performance on nuScenes test dataset}
% 我们还使用SOTA 3D检测器CenterPoint[79]和VoxelNeXt[80]作为基线对更大规模的nuScenes[13]数据集进行了实验，进一步验证了VoxelNextFusion的有效性，如表III所示。基于voxelnext，我们的SparseDet实现了66.8%的mAP和69.5%的NDS，比基线高出2.2%的mAP和1.3%的NDS。值得注意的是，在大多数的类别上我们的方法达到了Sota的Ap，其中的truck Barrier 以及 T.C. 我们的方法在 AP 中分别获得了 3.0%, 2.6% 和 3.5%的显着提高。 它充分证明了我们方法的有效性和有效性。总体而言，我们的方法改进了CenterPoint[79]和VoxelNeXt[80]的主要指标，导致mAP分别提高了8.8%和4.3%。得益于我们的LMFA模块的GFA模块，SparseDet可以充分聚合局部点云细节以及全局上下文信息有效的解决了稀疏代理特征表达能力不足的问题，故不仅允许对 “Motor.”、“Bike”、“Ped.”和“T.C”等小物体进行显着的性能改进，同时得益于充分的信息聚合，大型物体类别同样具备显著的性能改进。

We conduct experiments on larger-scale nuScenes~\cite{nuscenes} dataset using the SOTA (state-of-the-art) fully Sparse 3D detector VoxelNeXt~\cite{voxelnext} (a typical method of the second category mentioned in related works) as the baseline to further validate the effectiveness of our \textcolor{black}{SparseDet}, as shown in Table~\ref{tab_nuScens_test}. In the table, we not only compare our SparseDet with classical sparse detectors of the first category including FSD~\cite{FSD} and FSDV2~\cite{FSDV2}, but also include some SOTA dense 3D detectors such as UVTR~\cite{uvtr}, Transfusion~\cite{transfusion}, LargeKernel3D~\cite{largekernel3d} and PVT-SSD~\cite{pvt} for comparison.

% 出了sparse ,我们还比较了dense的方法
Based on VoxelNeXt~\cite{voxelnext}, our \textcolor{black}{SparseDet} achieves 66.7\% mAP and 71.3\% NDS, which surpasses the baseline by 2.2\% mAP and 1.3\% NDS.
It is worth noting that our method achieves state-of-the-art AP (Average Precision) on most categories. Specifically, for the categories of ``Truck", ``Barrier", and ``T.C.", our method shows significant improvements with AP gains of 3.0\%, 2.6\%, and 3.5\% respectively. 
Overall, the above results fully demonstrates the generalization and effectiveness of our method.
Thanks to our LMFA and GFA modules, SparseDet is able to obtain richer and more accurate object representations by comprehensively aggregating local point cloud details and global contextual information, while also facilitating the collaboration between the entire scene and instance features. As a result, it achieves significant performance improvements across a wide range of object categories.

\subsubsection{Performance on nuScenes val dataset}
% 为了证明我们的VoxelNextFusion框架的有效性，使用voxelnext [79]基线在nuScenes验证数据集上进行了实验。如表 IV 所示，我们的方法在“C”上优于 CenterPoint 11.6%、17.9% 和 18.0%。V.”、“Motor”和“Bike”类别。此外，我们的方法在不同大小的数据集中包含大量大物体的“Truck“和“Bus”类别上显示出显著的改进。这些结果表明由于充分的信息聚合有效的提升了稀疏物体检测代理的信息表达能力，同时由于将稀疏查询设计为物体检测代理，避免了中心特征缺失的问题、cite{SAFDNet}，上述结果表明了SparseDet在检测小目标以及大目标方面的有效性。
To demonstrate the effectiveness of our SparseDet, we conduct experiments on nuScenes validation dataset using VoxelNeXt\cite{voxelnext} as the baseline.
As shown in Table~\ref{tab_nuScenes_val}, our method outperforms VoxelNeXt by 6.4\%, 6.7\%, and 7.3\% on ``C.V.", ``Motor", and ``Bike" categories in the nuScenes full validation dataset, respectively. The results demonstrate that the effective aggregation of point cloud contextual information significantly enhances the representational capacity of the sparse object detection proxies. Additionally, by designing sparse queries as object detection proxies, we avoid the issue of missing central features, as mentioned in SAFDNet\cite{SAFDNet}. The aforementioned results demonstrate the effectiveness of SparseDet in detecting both small and large objects.

\subsubsection{Performance on KITTI test set}
 % 如表I所示，我们将VoxelNextFusion与KITTI测试集上的SOTA方法进行了比较。我们注意到，我们的VoxelNextFusion在3D和BEV检测的三个难度级别(3D AP为90.90%、82.93%、80.62%、BEV AP为94.46%、90.73%、88.34%)下表现出了出色的性能。为了公平比较，我们分别将体素 R-CNN [60] 和 PV-RCNN [61] 再现为强基线。值得注意的是，我们重新实现的结果几乎与[60]和[61]中报告的结果相同。我们的VoxelFusoin在大多数指标上都超过了Voxel R-CNN[60]。特别是在具有挑战性的硬水平上，我们分别在汽车、行人和骑自行车的人类别中分别提高了3.2%、3.02%和1.09%。同样，与 PV-RCNN [61] 相比，我们的方法在简单和中等级别上仅略有提高，而在硬级别上，我们大大超过了基线。与多模态方法Focals Conv[5]相比，我们的方法在汽车AP 3D的三个级别上分别实现了卓越的性能，分别提高了0.45%、0.65%和3.03%。总体而言，我们的VoxelNextFusion在KITTI[1]测试集上表现良好。特别是在主要由远处和小物体组成的硬级别上，这有力地证明了我们方法的有效性
 As shown in Table~\ref{tab_kitti_test}, we compare \textcolor{black}{SparseDet} with the SOTA methods on KITTI test set. We note that our \textcolor{black}{SparseDet} shows outstanding performance at three difficulty levels of 3D and BEV detection (90.79\%, 81.17\%, 78.11\% in 3D APs and 94.48\%, 89.68\%, 87.53\% in BEV APs). It is worth noting that VoxelNeXt~\cite{voxelnext} have not provided the results of KITTI dataset. For fair comparison, we reproduce VoxelNeXt~\cite{voxelnext} as the baseline, following the configuration exactly as described in the original paper\cite{voxelnext}. By Table~\ref{tab_kitti_test}, Our SparseDet surpasses VoxelNeXt~\cite{voxelrcnn} on most metrics. Especially, on the challenging hard level tasks, we improve 1.12\%, 2.72\% and 2.98\% in car, pedestrian, and cyclist categories, respectively. Compared to the latest dense approach, PVT-SSD\cite{pvt}, our method achieves superior performance, reaching 76.81\% and 63.46\% on the difficult levels for the car and two-wheeler categories, respectively. Overall, our \textcolor{black}{SparseDet} performs well on KITTI~\cite{kitti} test set, especially on the hard level which mostly consists of distant and small objects, strongly demonstrating the effectiveness of our method.

 \subsubsection{Performance on KITTI val dataset}
We further provide the results of the KITTI validation set to better present the detection performance of our \textcolor{black}{SparseDet}, as shown in Table~\ref{tab_kitti_val}. There are significant improvements compared to the baseline VoxelNeXt~\cite{voxelnext} on moderate and hard levels. Meanwhile, compared to dense detection approaches, our method also demonstrates competitive performance. For sparse detectors\cite{voxelnext} that treat a single center feature as a physical detection proxy, insufficient aggregation contextual information weakens the ability to present objects at central points, thus limiting the performance of the detection methods. The key factor that makes \textcolor{black}{SparseDet} effective is its ability to comprehensively aggregate contextual information at critical locations, thereby enhancing the object proxys' capacity to present objects.

\subsubsection{Comparison with Other Sparse Detectors}
% 为了证明我们的SparseDet的有效性，我们在nuS数据集上和其他的稀疏检测框架进行了对比。我们将SAFFDNet的推理速度与之前的性能最好的方法进行了比较，如表5所示。SparseDet 在达到最高mAP的同时仍然保持了优秀的推理速度。相比baseline voxelnext,在mAP方面实现2.2\%性能增益的同时，参数量仅仅增加了0.36M 同时由于不需要额外的任务进行辅助，在推理速度，模型参数量以及训练占用内存方面要优于先前的FSD和FSDV2
To demonstrate the effectiveness of our SparseDet, we conduct a comparative evaluation on nuScenes~\cite{nuscenes} dataset by comparing SparseDet against other sparse detection frameworks~\cite{voxelnext,FSD,FSDV2} as shown in Table~\ref{tab_sparse_compare}. In the table, we also compare the inference speed, the size of parameters and memory usage of SparseDet with prior sparse detection frameworks besides accuracy measurements. SparseDet achieves the highest mAP, meanwhile maintains an excellent inference speed. Compared to the baseline VoxelNeXt, SparseDet achieves a 2.2\% increase in mAP performance, but only has an increase of 0.36M parameters. Additionally, SparseDet does not need any auxiliary task. This is the main reason that SparseDet outperforms previous FSD\cite{FSD} and FSDV2\cite{FSDV2} methods in inference speed, the size of parameters, and training memory consumption.  
\subsection{Ablation Study}
\subsubsection{Effect of LMFA and GFA modules }

This section discusses the results of ablation experiments conducted on the baseline detector VoxelNeXt\cite{voxelnext} to evaluate the performance of each component in \textcolor{black}{SparseDet}. The results are reported in Table \ref{tab_ablation_kitti_voxel} and Table \ref{tab_ablation_nuscens} for KITTI and nuScenes $\frac{1}{4}$ subset, respectively.
Table \ref{tab_ablation_kitti_voxel} shows the initial AP scores for both AP$_{3D}$ and AP$_{BEV}$ on KITTI, which are 78.44\% and 87.10\%, respectively. As shown in Table \ref{tab_ablation_kitti_voxel}, LMFA and GFA modules lead to a significant improvement in performance on the hard-level KITTI tasks, with an increase of 4.27\% and 3.35\% for AP$_{3D}$ and AP$_{BEV}$, respectively. All the improvements do not significantly increase the model's parameter or weaken the inference speed.

% 如表\ref｛tab_ablation_nuscens｝所示，当使用LMFA sub-module时，SparseDet有一个极好的性能改进，这表明充分聚合上下文信息可以更好地增强稀疏特征的表达能力从而来增强稀疏3D检测器的性能。此外，当与GFA 集成时，增强作用进一步增强，mAP和NDS分别提高2.4%和1.3%。相比于旨在局部范围内增强关键位置特征的LMFA，GFA动态的从全局稀疏特征中学习并将信息聚合到物体检测代理中。得益于LMFA和GFA良好的上下文信息聚合能力，物体检测代理的表达能力得到极大的增强。值得注意的是，与KITTI相比，我们的\textcolor｛black｝｛VoxelNextFusion｝在大规模nuScenes数据集上同样产生了显著的改进。
As shown in Table \ref{tab_ablation_nuscens}, when using LMFA module, SparseDet achieves an excellent performance improvement, which indicates that effectively aggregating contextual information can better enhance the representation ability of sparse features, thereby improving the performance of the sparse 3D object detector. Compared to LMFA module, which focuses on learning local details of point clouds, the GFA module dynamically learns information from global sparse features. %and aggregates it into the object detection proxies. 
This promotes collaboration between scene and instance features, resulting in richer and more accurate object representations. When LMFA and GFA are integrated, this enhancement effect is further amplified, leading to improvements of 2.4\% in mAP and 1.3\% in NDS. %In general, by leveraging the powerful contextual information learning capabilities of LMFA and GFA, SparseDet greatly enhances the information representation ability of object proxies, resulting in significant performance advancements. 
% 研究结果强调了稀疏检测框架中上下文信息聚合的重要性，并为设计有效的聚合策略提供了宝贵的见解 
In summary, our ablation experiments show that \textcolor{black}{SparseDet} effectively enhances the performance of the baseline on challenging datasets. The research results underscore the importance of contextual information aggregation in sparse detection frameworks and provide valuable insights for designing effective aggregation strategies. %The results emphasize the significance of addressing the resolution gap between point clouds and images and offer valuable insights for designing effective fusion strategies. 

\begin{table}[t]
\scriptsize
\centering
\caption{Performance of different distances. The results are evaluated with AP calculated by 40 recall positions and 0.7 IoU threshold for \textcolor{black}{car category} in the \textcolor{red}{hard} level on KITTI \textit{val} set.
}
\renewcommand\arraystretch{1}
\setlength{\tabcolsep}{0.65mm}{
%\begin{tabular*}{\linewidth}{@{\extracolsep{\fill}}c|c|c|c|c|c|c}
%\begin{tabular*}{\linewidth}{@{}@{\extracolsep{\fill}}c|c|c|c|c|c|c@{}}

\begin{tabular}{l|c|c|c|c|c|c}

\toprule
 \multirow{3}{*}{Method}      & \multicolumn{3}{c|}{AP$_{3D}$(\%)}                                                            & \multicolumn{3}{c}{AP$_{BEV}$(\%)}                                      \\ \cmidrule(lr){2-7} 
                                                                     & 0-20m & 20-40m & 40m-inf & 0-20m & 20-40m & 40m-inf \\ \midrule
    
                                      VoxelNeXt$^*$        & 94.70 & 77.81  & 35.49   & 95.25 & 87.27  & 54.23
                                              \\
                                    \rowcolor{blue!10}\textcolor{black}{SparseDet}      & 95.72\textit{\fontsize{6}{0}\selectfont\textcolor{red}{+1.02}} & 79.95\textit{\fontsize{6}{0}\selectfont\textcolor{red}{+2.14}}  & 44.77\textit{\fontsize{6}{0}\selectfont\textcolor{red}{+9.28}}   & 96.66\textit{\fontsize{6}{0}\selectfont\textcolor{red}{+1.41}} & 89.26\textit{\fontsize{6}{0}\selectfont\textcolor{red}{+1.99}}  & 63.63\textit{\fontsize{6}{0}\selectfont\textcolor{red}{+9.40}}
                                              \\
\bottomrule
\end{tabular}
\label{tab_kitti_distances}
\begin{tablenotes}
\footnotesize
\item[1] * denotes re-implement result.
\item[1] The color \textcolor{red}{red} indicates improvement to our \textcolor{black}{SparseDet}.
\end{tablenotes}
}
\end{table}

\begin{table}[h!]
\scriptsize
\centering
\caption{Performance of different distances. The results are evaluated with mAP and NDS on nuScenes \textit{val} set.
}
\renewcommand\arraystretch{1}
\setlength{\tabcolsep}{1.65mm}{
%\begin{tabular*}{\linewidth}{@{\extracolsep{\fill}}c|c|c|c|c|c|c}
%\begin{tabular*}{\linewidth}{@{}@{\extracolsep{\fill}}c|c|c|c|c|c|c@{}}

\begin{tabular}{l|c|c|c|c|c|c}

\toprule
 \multirow{3}{*}{Method}      & \multicolumn{3}{c|}{mAP(\%)}                                                            & \multicolumn{3}{c}{NDS(\%)}                                      \\ \cmidrule(lr){2-7} 
                                                                     & 0-20m & 20-40m & 40m-inf & 0-20m & 20-40m & 40m-inf \\ \midrule
    
                                      VoxelNeXt$^*$        & 75.9 & 49.5  & 22.3   & 76.3 & 60.2  & 42.2
                                              \\
                                    \rowcolor{blue!10}\textcolor{black}{SparseDet}      & 77.6\textit{\fontsize{6}{0}\selectfont\textcolor{red}{+1.7}} & 55.3\textit{\fontsize{6}{0}\selectfont\textcolor{red}{+5.8}}  & 26.4\textit{\fontsize{6}{0}\selectfont\textcolor{red}{+4.1}}   & 77.3\textit{\fontsize{6}{0}\selectfont\textcolor{red}{+1.0}} & 63.8\textit{\fontsize{6}{0}\selectfont\textcolor{red}{+3.6}}  & 45.6\textit{\fontsize{6}{0}\selectfont\textcolor{red}{+3.6}}
                                              \\
\bottomrule
\end{tabular}
\label{tab_nus_distances}
\begin{tablenotes}
\footnotesize
\item[1] * denotes re-implement result.
\item[1] The color \textcolor{red}{red} indicates improvement to our \textcolor{black}{SparseDet}.
\end{tablenotes}
}
\end{table}

\subsubsection{Effect of the number of $M$}
The selection of neighboring voxel features, to enhance the feature representation at key locations, is a critical component of LMFA module. In this section, we will discuss the choice of the number of neighboring voxels $M$, and its corresponding effectiveness. Therefore, we configure different values for the hyperparameter $M$ (the number of neighboring voxels) including 4, 8, 16, and 32. As shown in Table~\ref{tab_ablation_nus_M}, the variation in the value of $M$ does not exhibit significant impact on the model's performance. It is worth noting that when $M$ is set to 8, our SparseDet model achieves the highest mAP, while setting $M$ to 16 results in the best NDS performance. Considering the overall model performance, inference time, training memory, and model parameters, we ultimately set $M$ to 8 as the default value.

\subsubsection{Effect of the number of $N_{key}$}
% 如表IX所示, 在nuS验证集上, 我们对LMFA模块中的关键体素数量N_key进行了消融研究。我们为超参数N_{key}配置了不同的值：500、1000、1500和2000。总的来说，随着N_key的值增加，SparseDet的模型表现对应的取得了不同程度的性能增益。另一个现象是模型性能受N_key值变化影响不大。这表明我们的SparseDet对超参数不是高度敏感的。尽管简单的增加N_key值有助于提升模型的表现，然而这是以模型的推理速度为代价的。综合考虑模型的精度以及推理延迟，我们选择500作为N_{key}的默认值
As shown in Table~\ref{tab_ablation_nus_N_Key}, we conduct an ablation study on the number of key voxels $N_{key}$ within LMFA module on the nuScenes validation dataset. We configure the value of the hyperparameter $N_{key}$ among 500, 1000, 1500, and 2000. In summary, as the value of $N_{key}$ increases, the performance of the SparseDet correspondingly achieves varying degrees of improvements. Another observation from the Table is that the model's performance does not exhibit a strong sensitivity to the changes of $N_{key}$. Although simply increasing the value of $N_{key}$ can enhance the model's performance, it comes at the cost of reduced inference speed.  After weighing the model's accuracy and inference latency, we ultimately select 500 as the default value for $N_{key}$.

\subsubsection{Effect of the number of $N_{K,V}$}
% 如表IX所示, 在nuS验证集上, 我们对GFA模块中的超参数N_key进行了消融研究。我们为超参数N_{key}配置了不同的值：6000、8000、10000和12000。总的来说，随着N_key的值增加，SparseDet的模型表现对应的取得了不同程度的性能增益。另一个现象是模型性能受N_key值变化影响不大。这表明我们的SparseDet对超参数不是高度敏感的。尽管简单的增加N_key值有助于提升模型的表现，然而这是以模型的推理速度为代价的。综合考虑模型的精度以及推理延迟，我们选择500作为N_{key}的默认值
%\textcolor{black}{
%As shown in Table \ref{tab_ablation_KITTI_T}, we conducted an ablation study on the crucial threshold $\mathcal{T}$ on the KITTI validation set. The range of $\mathcal{T}$ ranged from 0.1 to 0.9. Overall, when $\mathcal{T}$ is 0.5, our VoxelNextFusion achieved the better performance, and the performance variations were not substantial. It indicating that our VoxelNextFusion is not highly sensitive to hyperparameters.
%}
% 值得注意的是，当Nk,v取12000时SparseDet取得了最高的mAP和NDS,但同时有着最低的推理速度。综合考虑模型的精度以及推理延迟，我们选择500作为N_{key}的默认值。
As shown in Table~\ref{tab_ablation_nus_NKV}, we conduct an ablation study on the hyperparameter $N_{K,V}$ in the GFA module on the nuScenes validation set. We configure the value of the hyperparameter $N_{K,V}$ among 6000, 8000, 10000, and 12000. It is worth noting that when the value of $N_{K,V}$ is set to 12000, SparseDet achieves the highest mAP and NDS scores, but it exhibits the lowest inference speed. After weighing the model's accuracy and inference latency, we ultimately set $N_{K,V}$ to 10000 as the default value.

% 相比密集检测器，稀疏检测器的一大优势是可以将模型扩展到远端检测中而不增加太多的推理延迟。因此稀疏检测器能否稳定检测远端目标是衡量稀疏检测器性能的关键。在为了更好地了解我们的\textcolor｛black｝｛VoxelNextFusion｝在长距离下的卓越性能，我们在表\ref｛tab_dinstances｝中提供了不同距离范围的性能指标，特别是当硬级别包括更多的小对象和遮挡对象时。

% 这些结果清楚地反映了我们的\textcolor｛black｝｛SparseDet｝在较长距离上的优势，得益于SparseDet良好的对不同粒度的上下文学习能力显著增强了物体检测代理的特征表达能力，从而显著提高远处物体的精度。

\subsubsection{The performance of model in different distances}
Compared to dense detectors, a key advantage of sparse detectors is their ability to extend the models' long-range detection capabilities without a significant increase in inference latency. Consequently, the stable detection of distant targets is a critical metric for evaluating the performance of sparse detectors.To better understand the superior performance of our \textcolor{black}{SparseDet} at long distances, we provide performance metrics for different distance ranges in Table \ref{tab_kitti_distances} and Table \ref{tab_nus_distances}. Specifically, compared to the VoxelNeXt\cite{voxelnext}, our metrics show a more significant improvement, especially in the distance ranges of 20-40m and 40m-inf. For example, in 3D detection at 40m-inf of KITTI, our \textcolor{black}{SparseDet} improves $AP_{3D}$ by 9.28\%. In BEV detection at 40m-inf, our \textcolor{black}{SparseDet} improved $AP_{BEV}$ by 9.40\%. And on nuScenes dataset, in the detection at 40m~inf, our SparseDet improves by 4.1\% and 3.6\% on mAP and NDS, respectively. These results clearly reflect the advantages of our \textcolor{black}{SparseDet} model in longer-range detection. %Thanks to strong learning abilities for multi-scale contextual information of SparseDet, it significantly enhances the feature expression capability of the object detection model, which in turn significantly improves the accuracy of detecting distant objects.

% 这可以归因于这样一个事实，即我们的\textcolor｛black｝｛VoxelNextFusion｝更合理地利用了图像域中的语义信息，而不损害其语义和几何连续性的优势，这对于在难以建立几何关系的长程稀疏点云中利用成像的好处往往至关重要。总的来说，我们的方法在远程目标检测的精度上有了显著的提高
\begin{figure}[htb]
\centering \includegraphics[width=1\linewidth]{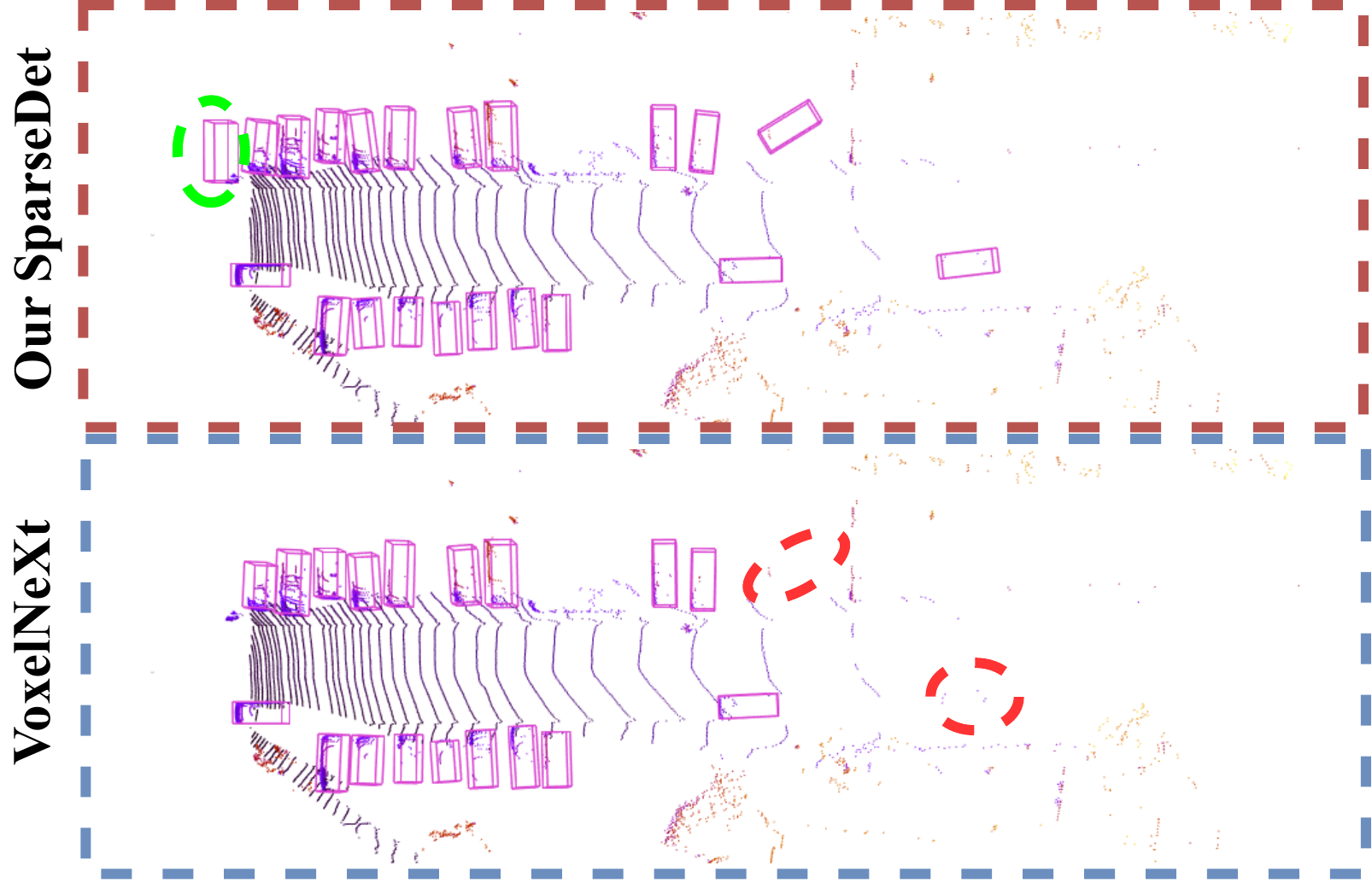}
 \caption{Visualization comparison between VoxelNeXt and our \textcolor{black}{SparseDet} in long-range detection, where false positives and false negatives are highlighted in \textcolor{green}{green} and \textcolor{red}{red}, respectively. 
 }
\label{fig:Visualization}
\end{figure}

\subsection{Visualization}
In Fig.~\ref{fig:Visualization}, compared to VoxelNeXt, we illustrate the superiority of our \textcolor{black}{SparseDet} for long-range/remote object detection, using the detection range of 0-70.4m for car category in KITTI as a case. According to the figure, our \textcolor{black}{SparseDet} has a false positive result, but there is no missed instance. Whereas, VoxelNeXt\cite{voxelnext} suffers from the missing of remote objects. This can be attributed to the fact that Our \textcolor{black}{SparseDet} fully utilizes the multi-scale contextual semantic information in point clouds, which are crucial for remote objects in sparse point clouds since these objects more obviously suffer from the lack of information. Overall, our method exhibits a significant improvement in the precision of remote object detection.

\section{Conclusions}
% 在这项工作中，我们提出了\textcolor｛black｝｛VoxelNextFusion｝，这是一种用于多模式3D对象检测的简单、统一和有效的体素融合框架。具体而言，我们基于四种经典的基于体素的方法，即体素R-CNN、PV-RCNN、CenterPoint和VoxelNeXt，设计了一个统一的多模态框架，更合理地利用了图像语义信息和背景信息，从而增强了泛化和鲁棒性。综合实验结果表明，\textcolor｛black｝｛VoxelNextFusion｝显著提高了3D检测器在KITTI和nuScenes数据集上的性能。我们希望我们的工作能够为自动驾驶的多模态特征融合提供新的见解
In this work, we propose \textcolor{black}{SparseDet}, a simple and effective framework for fully sparse 3D object detection. Specifically, based on VoxelNeXt, we have designed an efficient sparse detection framework that more reasonably utilizes instance-level and scene-level point cloud contextual information. This has significantly enhanced the expression capability of object proxies, thereby substantially improving the detection performance of a sparse detector. Comprehensive experimental results demonstrate that \textcolor{black}{SparseDet} significantly improves the performance compared with the baseline on the KITTI and nuScenes datasets. We hope our work can provide new insights into sparse detectors for autonomous driving. 

Currently, the research works on sparse 3D detectors are not sufficient as the other directions such as multi-modal 3D detection. This makes the comparison methods of 3D sparse frameworks be limited. %What's more, our computational resources do not support our experiments on more larger datasets such as Waymo. 
However, for real-world applications, the latency of a model is very important. Therefore, research on fully sparse and fast detectors deserves more attention and focus.

\addtolength{\textheight}{-0cm}


%\bibliographystyle{piain}
% \bibliography{biblio/IROS}

% IEEEtran.bst
\bibliographystyle{IEEEtran}
\bibliography{IEEEabrv, main}
%IEEEabrv instead of IEEEfull
%\begin{thebibliography}{00}
%
%%% \bibitem[Author(year)]{label}
%%% Text of bibliographic item
%\end{thebibliography}

\begin{IEEEbiography}[{\includegraphics[width=1in,height=1.25in,clip,keepaspectratio]{{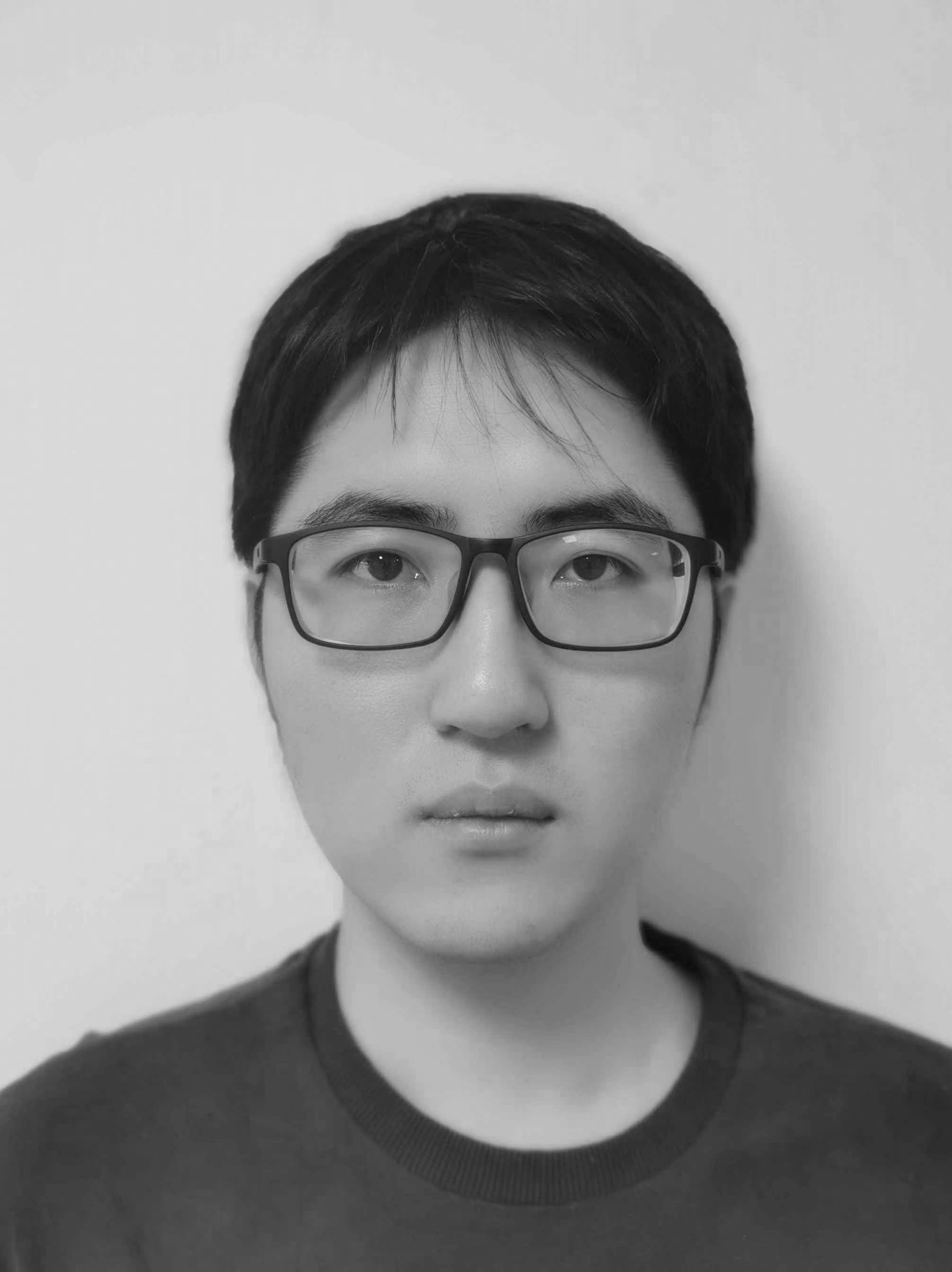}}}]{Lin Liu} was born in Jinzhou, Liaoning Province, China, in 2001. he received his bachelor’s degree from China University of Geosciences(Beijing). Now, he is studying for his master's degree at the Beijing Jiaotong University (China). His research interests are in computer vision.%Since December 2022, he has worked as an intern in the School of Computer and Information Technology of Beijing Jiaotong University. At present, he has been recommended to be a master's student of computer science and technology at Beijing Jiaotong University without examination.

\end{IEEEbiography}

\begin{IEEEbiography}[{\includegraphics[width=1in,height=1.25in,clip,keepaspectratio]{{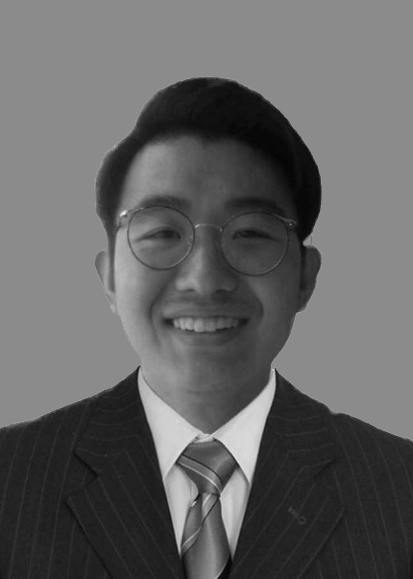}}}]{Ziying Song} was born in Xingtai, Hebei Province, China, in 1997. He received his B.S. degree from Hebei Normal University of Science and Technology (China) in 2019. He received a master's degree from Hebei University of Science and Technology (China) in 2022. He is now a Ph.D. student majoring in Computer Science and Technology at Beijing Jiaotong University (China), with research focus on Computer Vision.

\end{IEEEbiography}

\begin{IEEEbiography}[{\includegraphics[width=1in,height=1.25in,clip,keepaspectratio]{{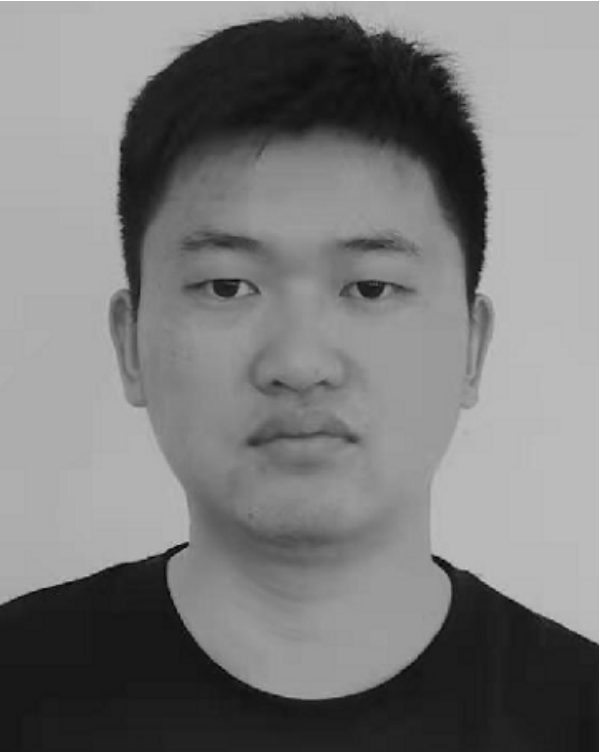}}}]{Qiming Xia} is currently pursuing the master’s
degree with the Computer Engineering College,
Jimei University, Xiamen, China, in 2013.

His research interests include 3-D object detection
of point clouds, machine learning, deep learning
theory and its application, knowledge graph, and
image processing.

\end{IEEEbiography}

\begin{IEEEbiography}
[{\includegraphics[width=1in,height=1.25in,clip,keepaspectratio]{{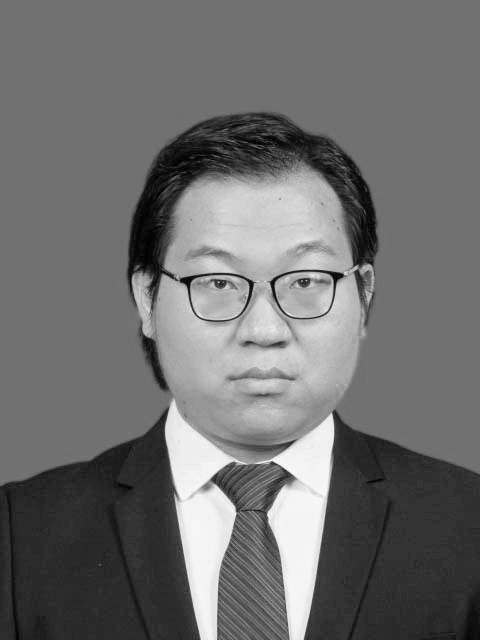}}}]
{Feiyang Jia} was born in Yinchuan, Ningxia Province, China, in 1998. He received his B.S. degree from Beijing Jiaotong University (China) in 2020. He received a master's degree from Beijing Technology and Business University (China) in 2023. He is now a Ph.D. student majoring in Computer Science and Technology at Beijing Jiaotong University (China), with research focus on Computer Vision.
\end{IEEEbiography}

\begin{IEEEbiography}[{\includegraphics[width=1in,height=1.25in,clip,keepaspectratio]{{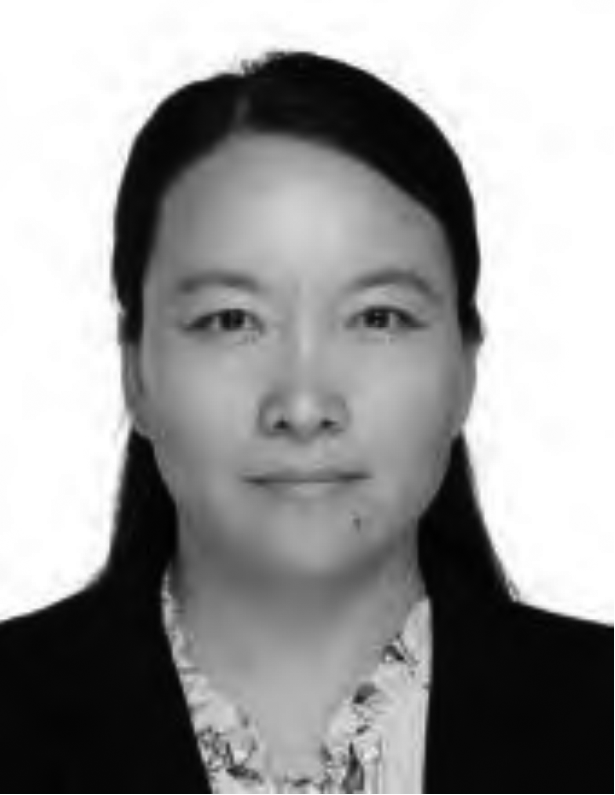}}}]{Caiyan Jia} was born in 1976. She received her Ph.D. degree from Institute of Computing Technology, Chinese Academy of Sciences, China, in 2004. She had been a postdoctor in Shanghai Key Lab of Intelligent Information Processing, Fudan University, Shanghai, China, in 2004–2007. She is now a professor in School of Computer and Information Technology, Beijing Jiaotong University, Beijing, China. Her current research interests include deep learning in computer vision, graph neural networks and social computing, etc.

\end{IEEEbiography}

\begin{IEEEbiography}[{\includegraphics[width=1in,height=1.25in,clip,keepaspectratio]{{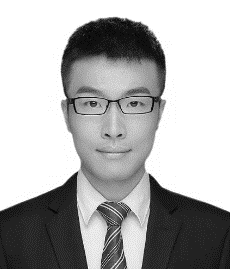}}}] {Lei Yang (Graduate Student Member, IEEE)} received his B.E. degree from Taiyuan University of Technology, Taiyuan, China,
and the M.S. degree from the Robotics Institute at Beihang University, in 2018. Then he joined the Autonomous Driving R$\&$D Department of JD.COM as an algorithm researcher from 2018 to 2020. He is now a Ph.D. student in the School of Vehicle and Mobility at Tsinghua University since 2020.
His current research interests include computer vision, 3D scene understanding and autonomous driving.

\end{IEEEbiography}

\begin{IEEEbiography}[{\includegraphics[width=1in,height=1.25in,clip,keepaspectratio]{{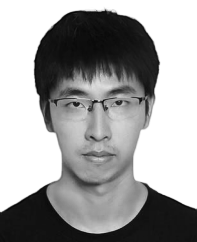}}}]{Hongyu Pan}  received the B.E. degree from Beijing Institute of Technology (BIT) in 2016 and the M.S. degree in computer science from the Institute of Computing Technology (ICT), University of Chinese Academy of Sciences (UCAS), in 2019. He is currently an employee at Horizon Robotics. His research interests include computer vision, pattern recognition, and image processing. He specifically focuses on 3D detection/segmentation/motion and depth estimation.

\end{IEEEbiography}

\end{document}